\PassOptionsToPackage{dvipsnames,table}{xcolor}

\documentclass[acmsmall,screen]{acmart}
\usepackage{booktabs} 
\usepackage{graphicx}
\usepackage[export]{adjustbox}
\usepackage{xcolor}

\usepackage{multirow}
\usepackage{amsmath}
\usepackage{xspace}
\usepackage{xcolor}
\usepackage{dsfont}
\usepackage{rotating}
\usepackage{makecell}
\usepackage[ruled,vlined]{algorithm2e}
\usepackage{amsthm}
\usepackage{subcaption}
\renewcommand{\epsilon}{\varepsilon}

\usepackage{mathtools}

\newcommand{\R}{\mathbb{R}}

\begin{document}

\title{Cascading CMA-ES Instances for Generating Input-diverse Solution Batches}
\author{Maria Laura Santoni}
\orcid{}
\affiliation{
  \institution{Sorbonne Universit\'e, CNRS, LIP6}
  \city{Paris}
  \country{France}}

\author{Christoph Dürr}
\orcid{0000-0001-8103-5333}
\affiliation{
  \institution{Sorbonne Universit\'e, CNRS, LIP6}
  \city{Paris}
  \country{France}}

\author{Carola Doerr}
\orcid{0000-0002-4981-3227}
\affiliation{
  \institution{Sorbonne Universit\'e, CNRS, LIP6}
  \city{Paris}
  \country{France}}

\author{Mike Preuss}
\orcid{0000-0003-4681-1346}
\affiliation{
  \institution{Universiteit Leiden}
  \city{Leiden}
  \country{The Netherlands}}

\author{Elena Raponi}
\orcid{0000-0001-6841-7409}
\affiliation{
  \institution{Universiteit Leiden}
  \city{Leiden}
  \country{The Netherlands}
}

\begin{abstract}

Rather than obtaining a single good solution for a given optimization problem, users often seek alternative design choices, because the best-found solution may perform poorly with respect to additional objectives or constraints that are difficult to capture into the modeling process.

Aiming for batches of diverse solutions of high quality is often desirable, as it provides flexibility to accommodate post-hoc user preferences. At the same time, it is crucial that the quality of the best solution found is not compromised.

One particular problem setting balancing high quality and diversity is fixing the required minimum distance between solutions while simultaneously obtaining the best possible fitness.
Recent work by Santoni et al. [arXiv 2024] revealed that this setting is not well addressed by state-of-the-art algorithms, performing in par or worse than pure random sampling. 

Driven by this important limitation, we propose a new approach, where parallel runs of the covariance matrix adaptation evolution strategy (CMA-ES) inherit tabu regions in a cascading fashion.   
We empirically demonstrate that our CMA-ES-Diversity Search (CMA-ES-DS) algorithm generates trajectories that allow to extract high-quality solution batches that respect a given minimum distance requirement, clearly outperforming those obtained from off-the-shelf random sampling, multi-modal optimization algorithms, and standard CMA-ES. 

\end{abstract}



\maketitle

\sloppy
\section{Introduction}

Unlike traditionally taught in basic optimization courses, where the focus is on identifying a single solution of best-possible quality, practical optimization problems often demand for a different perspective. In real-world optimization scenarios, returning only one global solution is often undesirable, as the problem modeling may neglect constraints or objectives that are hard to formalize.
Instead, providing batches of diverse alternative high-quality solutions is crucial, as they allow for post-hoc choices based on the compromises a decision-maker must consider when deviating from a single proposed solution. 

An archetypal example is engineering design, where certain manufacturing constraints and costs are challenging to model into the optimization problem. Thus, final expert judgment is needed before deciding which designs are kept for more advanced stages of the development process.  
In such settings, it is crucial to provide a batch of diverse 
solutions to designers~\citep{raponi_kriging-assisted_2019,dommaraju_identifying_2019,yousaf_similarity-driven_2023,bujny_learning_2023}. 
However, ensuring that the proposed solutions are not only diverse but also include the best possible option is critical to enhance the utility of such methods for users.

Similar questions have been investigated across various subfields of optimization, though from perspectives that differ from the focus of this paper. For instance, Evolutionary Diversity Optimization (EDO)~\citep{ronaldFindingMultipleSolutions1995,zechmanEvolutionaryAlgorithmGenerate2004} focuses on evolving sets of solutions that meet a fixed quality threshold while maximizing diversity. Quality Diversity (QD)~\citep{DBLP:journals/firai/PughSS16,DBLP:journals/tec/CullyD18} seeks to illuminate the solution space by exploring diverse niches in the feature space and maximizing quality within each niche. Similarly, Multimodal Optimization (MMO)~\citep{Preuss2021} aims to identify multiple global and local optima by exploring various modes in the search space. 
A recent study by Santoni et al. in~\citep{santoni2024illuminatingdiversityfitnesstradeoffblackbox} explores the trade-off between input-space diversity and quality in batches of equally weighted solutions. Their results highlight that the histories generated by default approaches, e.g. MMO algorithms, are often insufficient for extracting high-quality and diverse batches of solutions under a distance constraint in the input space, which highlights the need for targeted algorithm design for this optimization scenario. 

\begin{figure}[t] \center
    \includegraphics[width=\textwidth]{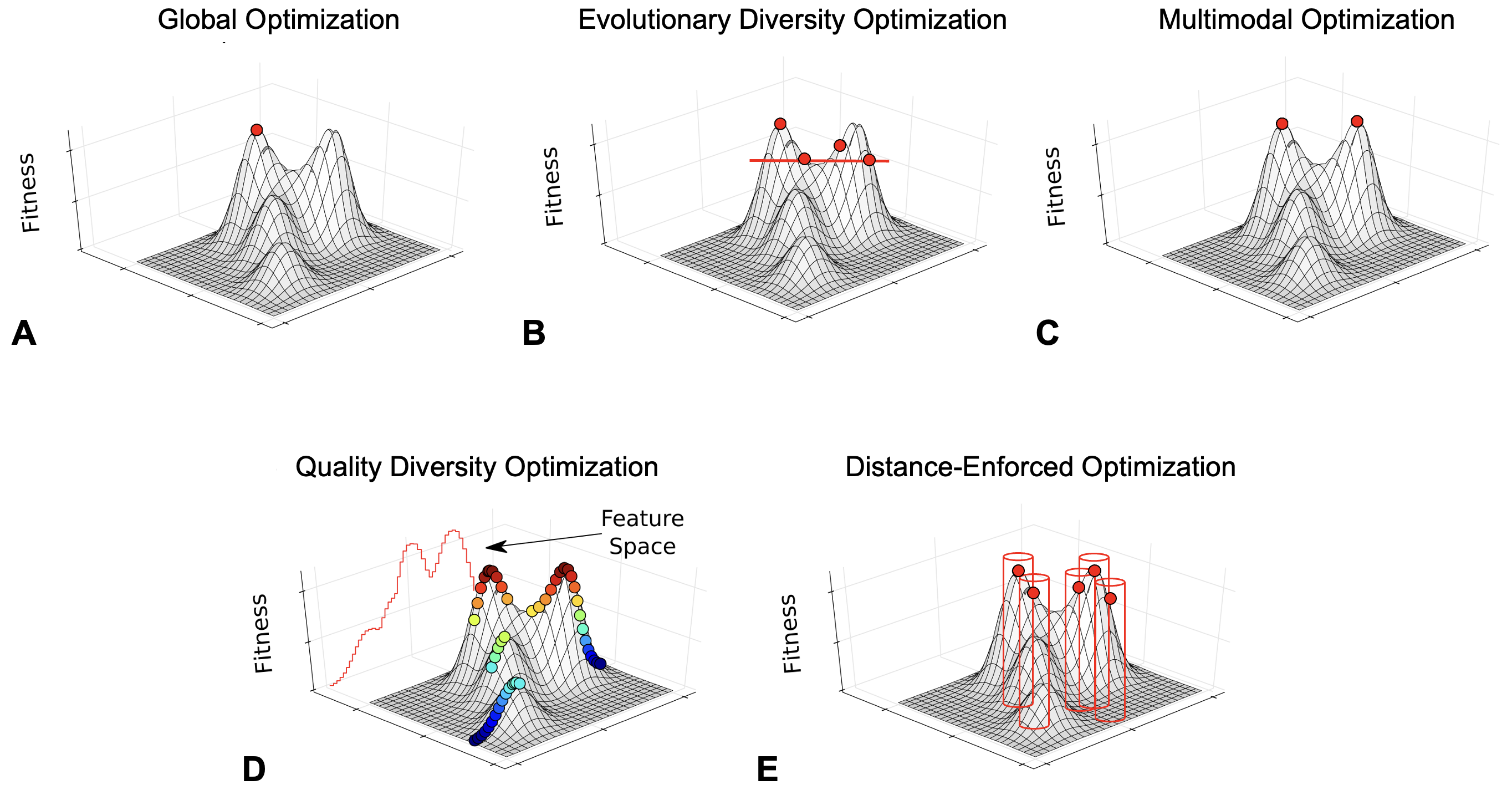}
      \caption{Conceptual overview of four optimization paradigms: Global optimization (A), evolutionary diversity optimization (B), multimodal optimization (C), quality diversity (D), and distance-enforced optimization (E), addressed in this work. Figure adapted from~\citep{vassiliades2017usingcentroidalvoronoitessellations}.}
    \label{fig:paradigms_new}
\end{figure}

Figure~\ref{fig:paradigms_new} illustrates the conceptual difference between established optimization paradigms and our target setting (D). While, unlike global optimization (A), methods such as EDO (B), MMO (C), QD (D) aim to explore sets of diverse, high-quality solutions, they typically operate under assumptions about modality, predefined feature spaces, or fixed quality thresholds.
These assumptions limit their ability to extract multiple high-performing solutions that are explicitly diverse in the input space without compromising the quality of the best found solution. Inspired by~\citep{santoni2024illuminatingdiversityfitnesstradeoffblackbox}, our work addresses a novel optimization scenario that seeks batches of input-space-diverse, high-quality solutions, as illustrated in (E).

\textbf{Our Contribution:} 

\begin{figure}[t] \center
    \includegraphics[trim=5.5cm 2.1cm 5.5cm 7.5cm, clip, width=.7\textwidth]{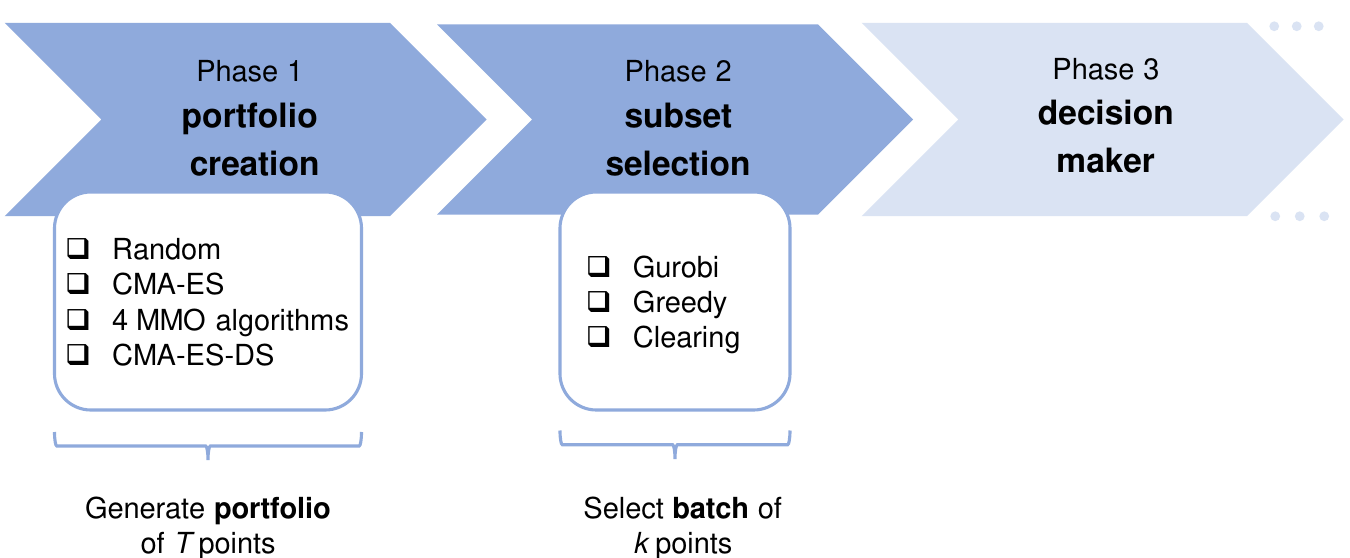}
      \caption{Three-phase optimization process: Phases 1 and 2 (covered in this paper) are in full color, while the final phase (beyond this work) is in a lighter shade.}
    \label{fig:process_phases}
\end{figure}

We address the diversity-enforced optimization challenge by assuming that the optimization process can be decomposed into three distinct phases, as illustrated in Figure~\ref{fig:process_phases}.
In Phase 1, an initial portfolio of $T$ points is created, through random sampling or as the trajectory of an optimization algorithm. In Phase 2, a batch of $k$ points is extracted from this portfolio. Finally, in Phase 3, the decision maker uses this batch to make further selection or adjustments to the proposed solutions in the batch, taking into account additional objectives and constraints. This latter phase is not further addressed in our work.

We focus primarily on the first phase, proposing a new algorithm based on the covariance matrix adaptation evolution strategy (CMA-ES)~\citep{hansen_reducing_2003}, which we extend by a cascading diversity mechanism to respect an imposed minimum distance requirement. 
Following the approach in~\citep{santoni2024illuminatingdiversityfitnesstradeoffblackbox}, we consider the Euclidean distance as the input diversity metric to enforce the minimum distance requirement. This choice provides a clear and practical proof of concept. However, our approach is flexible and can easily incorporate any problem-specific diversity metric, depending on the context of the application. 
Using these requirements, we compare the batches extracted from the trajectory generated by our CMA-ES-Diversity Search (CMA-ES-DS) algorithm against those obtained through MMO algorithms, random sampling, and standard CMA-ES. 

We evaluate three different strategies for extracting the batches from the full search trajectories (Phase 2):
(1) an exact but computationally expensive method based on the Gurobi solver~\citep{gurobi};
(2) a faster greedy heuristic proposed in~\citep{santoni2024illuminatingdiversityfitnesstradeoffblackbox}; and
(3) a simple \textit{clearing} approach that performs a single pass over the trajectory, greedily adding points that satisfy the minimum distance requirement with respect to the already selected ones, i.e., that are at least $d_{\min}$ apart from every point already included in the batch.
The third approach, inspired by the clearing scheme introduced by Petrowski in~\citep{542703}, has proven effective across various optimization applications where maintaining diversity and exploring multiple solutions is relevant~\citep{eriksson2019scalable, maus2023discoveringdiversesolutionsbayesian, WANG2022100976, 10.1007/978-3-642-30976-2_42, 5363222}.

We empirically evaluate CMA-ES-DS on the 24 BBOB functions~\citep{hansen2021coco} in dimensions $d\in\{2,5,10\}$, using a budget of $T\in\{1\,000, 3\,000, 5\,000, 10\,000\}$ function evaluations, batch sizes $k \in \{3,5,7\}$ and at least three different distance requirements per setting. These trajectories are compared to seven off-the-shelf methods (random sampling, two versions of default CMA-ES, and four multi-modal optimization algorithms). We show that our algorithm generates portfolios from which superior batches of solutions can be extracted across the diverse settings, with particularly pronounced improvements observed in higher dimensions and under low-budget constraints, where the trade-off between diversity and quality becomes more challenging.

\textbf{Note on prior publication:} A short version of this work was presented at a workshop with archived proceedings (names and reference are omitted here for double-blind review purposes and will be added for an eventual camera-ready paper).
 
In summary, this present version extends the workshop paper by the following:
\begin{itemize}
    \item Instead of six algorithms, we consider seven by including APDMMO~\citep{ma2025accuratepeakdetectionmultimodal}.
    \item Instead of considering only batch size $k = 5$, we additionally evaluate $k = 3$ and $k = 7$.
    \item Instead of using only budget $T =$ $1\,000$, and $T =10\,000$, we include intermediate values $3\,000$ and $5\,000$ for CMA-ES-DS.
    \item Instead of reporting only loss-based performance, we include a CPU time analysis across varying $T$ and $k$, revealing additional challenges in some settings (e.g., difficulties in placing points and generating valid populations).
    \item Instead of fixing the CMA-ES-DS variant, we conduct an ablation study across its versions, confirming that the chosen configuration is the most robust and reliable.
    \item Instead of a single instance per function, we analyze multiple instances—specifically, instances from 0 to 20 inclusive—to assess the performance variability of CMA-ES-DS.
\end{itemize}

\textbf{Reproducibility:} The code for reproducing our experiments, along with the whole set of figures are available on GitHub~\citep{code}.

\section{Problem Statement}
\label{sec:probstat}
\textbf{The general problem.} 
Building on the ideas presented in~\citep{santoni2024illuminatingdiversityfitnesstradeoffblackbox}, we adapt the general problem to address a slightly different objective. Specifically, we aim to construct a final batch of $k$ solutions that does not only satisfy a minimum pairwise distance constraint but also adheres to a lexicographic ordering of quality. In this ordering, the first solution is the best (ideally approximating the quality of the global optimum as well as possible), serving as a \textit{leader}, followed by $k-1$ alternative solutions. 

Given a (possibly unknown) function $f:S \subseteq \R^D \rightarrow \R$, a minimum distance requirement $d_{\min}>0$, and a batch size $k$, our problem is to find a collection $X^* = \{x^1, \ldots, x^k\} \subseteq S$ such that the pairwise (Euclidean) distance between any two points $x^i$ and $x^j$ with $i \neq j$ satisfies $d(x^i, x^j) \geq d_{\min}$ and such that: (i) the first point $x^1$ is the best possible solution with respect to $f$ (i.e., $x^1 = \arg\min_{x \in S} f(x)$), and (ii) the batch of $k$ suggested solutions $x^1, \ldots, x^k$ minimize the average objective value \( \sum_{i=1}^k f(x^i) / k \). 

We note that our problem could be interpreted as a multi-objective one, with the solution quality of the leader as first objective and that of the entire batch as second. However, we are not (or more precisely, not here in this work) interested in the classical multi-objective way of exploring the Pareto front formed by these two objectives. Instead, we prioritize the quality of the leader while ensuring diversity among the latter and the alternatives. %

\textbf{Constructing the batch of $k$ solutions.} As in~\citep{santoni2024illuminatingdiversityfitnesstradeoffblackbox}, we first construct initial portfolios $X$ containing $T$ evaluated points. These portfolios are either randomly sampled points or the full search trajectory of an iterative black-box optimization algorithm using a total budget of $T$ function evaluations. 
From these portfolios $X$, we then seek to identify a feasible batch $X^*=\{x^1,\ldots,x^k\} \subseteq X$ of $k$ solutions that are $d_{\min}$ far apart from another (more precisely, we require the minimal pairwise distance $\min_{1 \le i < j \le k}d(x^i,x^j)$ to be at least $d_{\min}$). 

To compare the different algorithms, we will not only consider the average quality $\sum_{i=1}^k f(x^i) / k$ of the selected solutions, but we rather consider the averages of the best $\ell$ solutions in the selected batch $X^*$, for all $1\le \ell \le k$. Note that this is similar to a pairwise comparison of the $\ell$-th best solutions of each batch as the value of the $\ell$-th best can be extracted from knowing the average of the $\ell-1$ best and the $\ell$ best.  

Since we do not want to compromise on the quality of the best solution returned, we always select $x^1 = \arg\min X$, ties broken arbitrarily (in our continuous setting, we do not encounter solutions with exactly the same objective value). The impact of this requirement depends on the function. For a unimodal function with steep slopes around the global optimum, forcing $x^1 = \arg\min X$ to be included in the batch $X^*$ can be costly, wheras the impact on a multimodal function with disperse local optima of similar quality will be negligible.  

\textbf{Disclaimer.} We note that none of the algorithms studied in this paper (i.e., neither the baselines nor our proposed CMA-ES variant) uses the rolling averages as an immediate optimization criterion during the search.

\section{Portfolio Creation via CMA-ES-DS}
\label{sec:portfoliocreation}

Our CMA-ES-Diversity Search (CMA-ES-DS) algorithm extends the Covariance Matrix Adaptation Evolution Strategy (CMA-ES)~\citep{hansen_reducing_2003} by incorporating mechanisms to promote diversity in the history of the evaluated points. Whereas CMA-ES is a powerful algorithm, it does not explicitly encourage diversity among the evaluated solutions, rendering it less useful for our optimization scenario~\citep{santoni2024illuminatingdiversityfitnesstradeoffblackbox}. 

\begin{figure*}[h!] \center
    \includegraphics[trim=0cm 0cm 0cm 0cm, clip,width=\textwidth]{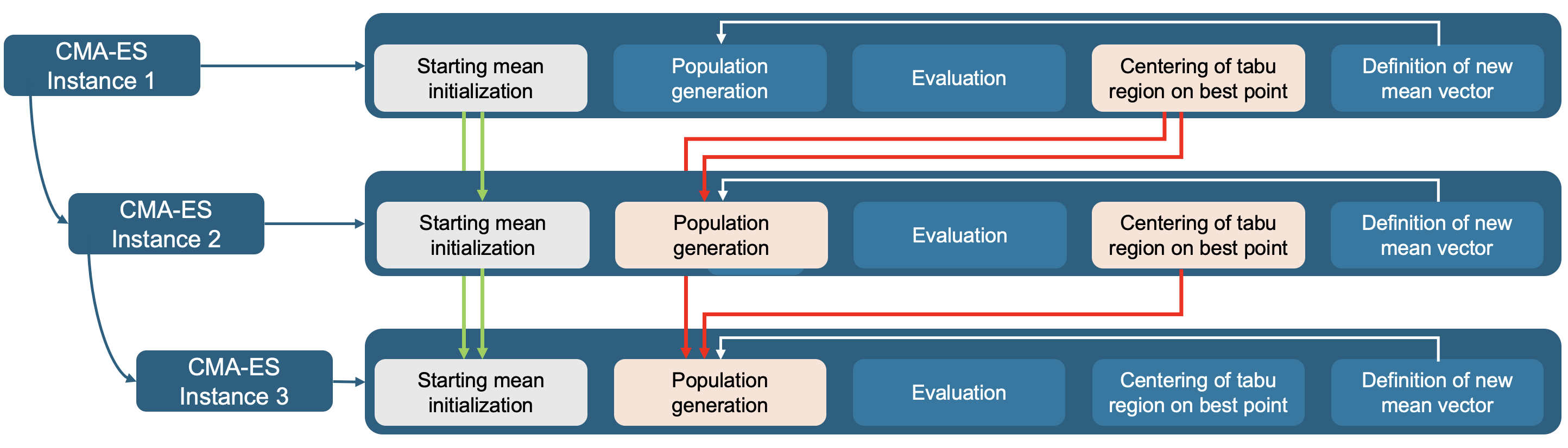}
      \caption{Cascade control mechanism in CMA-ES-DS with batch size $k = 3$. The diagram highlights the two key phases: the initialization of diverse starting means (grey boxes) and the synchronized generation of populations (pink boxes) for three CMA-ES instances. Green arrows highlight the dependency during initialization, where each mean must respect a minimum distance constraint ($d_{\min}$) from previously initialized means. Red arrows represent the cascading control during population generation, preventing later instances from generating candidates within the tabu regions of earlier ones.     
      }
    \label{fig:cascade}
\end{figure*}

CMA-ES-DS achieves diversity through a cascading control mechanism, which operates during two critical phases: the initialization of diverse starting means and the synchronized evolution of populations across multiple CMA-ES optimizers, each one referred to as an \textit{instance}. As illustrated in Figure ~\ref{fig:cascade}, the cascading influence ensures that the solutions generated by later instances respect the diversity constraints defined by the earlier ones. These mechanisms are implemented through cascading validation checks during initialization (grey boxes, green arrows in Figure ~\ref{fig:cascade}) and during population generation (pink boxes, red arrows in Figure ~\ref{fig:cascade}).

The pseudocode of the algorithm is presented in Algorithm~\ref{alg:CMAES-D-S}, which provides a high-level structure of the method. The algorithm operates with $k$ independent CMA-ES instances, each of which maintains a tabu region to ensure diversity. These instances evolve synchronously, updating their respective tabu regions dynamically.

\begin{algorithm}[t]
\caption{Pseudocode of the CMA-ES-DS algorithm}

\label{alg:CMAES-D-S}

\KwIn{
  $f$: objective function to minimize, 
  $d_{\min}$: minimum distance requirement, 
  $k$: nbr of CMA-ES instances ($i$-th one denoted as CMA-ES$_i$), 
  $T$: total budget of function evaluations
}

\BlankLine

\For{$i = 1 \dots k$}{
        Initialize $x_{0_i}$ respecting the minimum distance requirement with cascade check\;
        Initialize CMA-ES\textsubscript{i} starting from $x_{0_i}$\;
    }
\While{$nb\_fevals \leq T$}{
        \For{$i = 1 \dots k$}{
            \While{$nb_\_evaluated\_points < \lambda$}{
                Sample a point from the CMA-ES\textsubscript{i} distribution\;
                Check if the point is valid with cascade check:\\
                \If{point is valid}{Evaluate the point\;
                
                }
            
            }
        Update tabu region\;
        }
    }

\Return portfolio of $T$ points\;

\end{algorithm}

The key components of CMA-ES-DS are hence as follows.  

\textbf{Initialization of Diverse Starting Means:} The algorithm begins by generating an initial set of $k$ starting points ($x_{0_i}$), each serving as the initial mean of the distribution for an independent  CMA-ES instance. To ensure diversity, these means are guaranteed to be at least $d_{\min}$ apart one from another. The initialization follows a cascading approach: the first mean is generated freely, the second mean is constrained to respect the minimum distance from the first, the third mean considers both the first and the second, and so forth. This process is implemented using rejection sampling, where candidate means are drawn uniformly at random within the search space and are included only if they satisfy the distance constraint relative to previously selected means. This cascading mechanism is illustrated in Figure ~\ref{fig:cascade}, where the grey boxes represent the initialization of means, and the green arrows indicate the dependency and checks performed for each subsequent mean.
    
\textbf{Tabu Regions for Diversity Maintenance:} Each CMA-ES instance generates its own tabu region, defined as a sphere of a given radius ($d_{\min}$). Points generated by each CMA-ES instance must not fall within the tabu regions of the preceding instances, enforcing diversity across the search space.
    
\textbf{Dynamic Updates to Tabu Regions:} The tabu regions are initialized with the starting mean ($x_{0_i}$) of each CMA-ES instance as centers. At each iteration, the center is dynamically updated to the best point in terms of fitness value from the population of the respective CMA-ES instance. 

When a CMA-ES instance meets a stopping criterion, the center of its tabu region is set to the best point it has evaluated.

\textbf{Cascading Validation and Synchronized Evolution:} Each CMA-ES instance evolves its population in a synchronized manner, where all instances complete their first iteration before moving to the second, and so on. 
During the population generation phase, each CMA-ES instance samples candidate solutions from its multivariate normal distribution, rejecting those that fall within the tabu regions of preceding CMA-ES instances in the cascading order, until the desired population size $\lambda$ is reached.
The first CMA-ES instance operates without any distance constraint and works as a standard CMA-ES, which supports the requirement of having a leader by allowing an unconstrained search for high-quality solutions. 
For the following instances, a cascading control mechanism is applied to ensure that each candidate solution respects the tabu regions established by all preceding instances in the cascade. Specifically, candidates that fall within the tabu regions of any preceding CMA-ES instance in the cascade are rejected and resampled. 

This procedure, referred to as \textit{cascade check} in Algorithm~\ref{alg:CMAES-D-S}, guarantees that the points generated by later CMA-ES instances fall outside of the tabu regions of earlier instances.
In Figure ~\ref{fig:cascade}, the pink boxes and the red arrows highlight the influence of preceding tabu regions' instances on the population generation of subsequent ones.

\textbf{Restart Mechanism:} If all CMA-ES instances reach a stopping criterion before the total evaluation budget is exhausted, the algorithm restarts, reinitializing with a new set of diverse starting means. 
If fewer than all instances meet their stopping criteria, the remaining instances continue their optimization process until they either converge or the evaluation budget is used up. 
When one of the CMA-ES instances converges, its tabu region center is set to the best solution it has found so far for all the remaining iterations until a complete restart of the algorithm happens or budget is exhausted.

We chose to set the center of each tabu region to be the best point in the current population. This ensures that the centers remain tied to actual high-quality solutions while straightforwardly enforcing the diversity constraints. Other alternatives were tested, such as centering on the best-so-far solution from each CMA-ES instance or on the mean of the distribution, but both approaches yielded worse results. Ultimately, centering on a specific evaluated point from the population provided the best results.
A detailed comparison of these variants is provided in the Appendix.

Figure ~\ref{fig:algorithm_phases} illustrates the evolution of CMA-ES-DS over the isocontour of the f2 problem from the BBOB benchmark suite~\citep{hansen2021coco} (see Section ~\ref{sec:exset} below) in dimension 2, with a maximum of $1\,000$ function evaluations, batch size $k=5$, and distance constraint $d_{\min}=3$.
We see that the tabu centers progressively move towards regions with better function values, remain at least $d_{\min}$ apart throughout the run, and that, after clearing, the final batch solutions follow the cascade order—instance 1 lies closest to the global optimum, while instance 5 ends in a slightly worse isocontour region.

\begin{figure*}[h!] \center
   \includegraphics[trim=0cm 4.3cm 0cm 4.5cm, clip,width=0.82
    \textwidth]{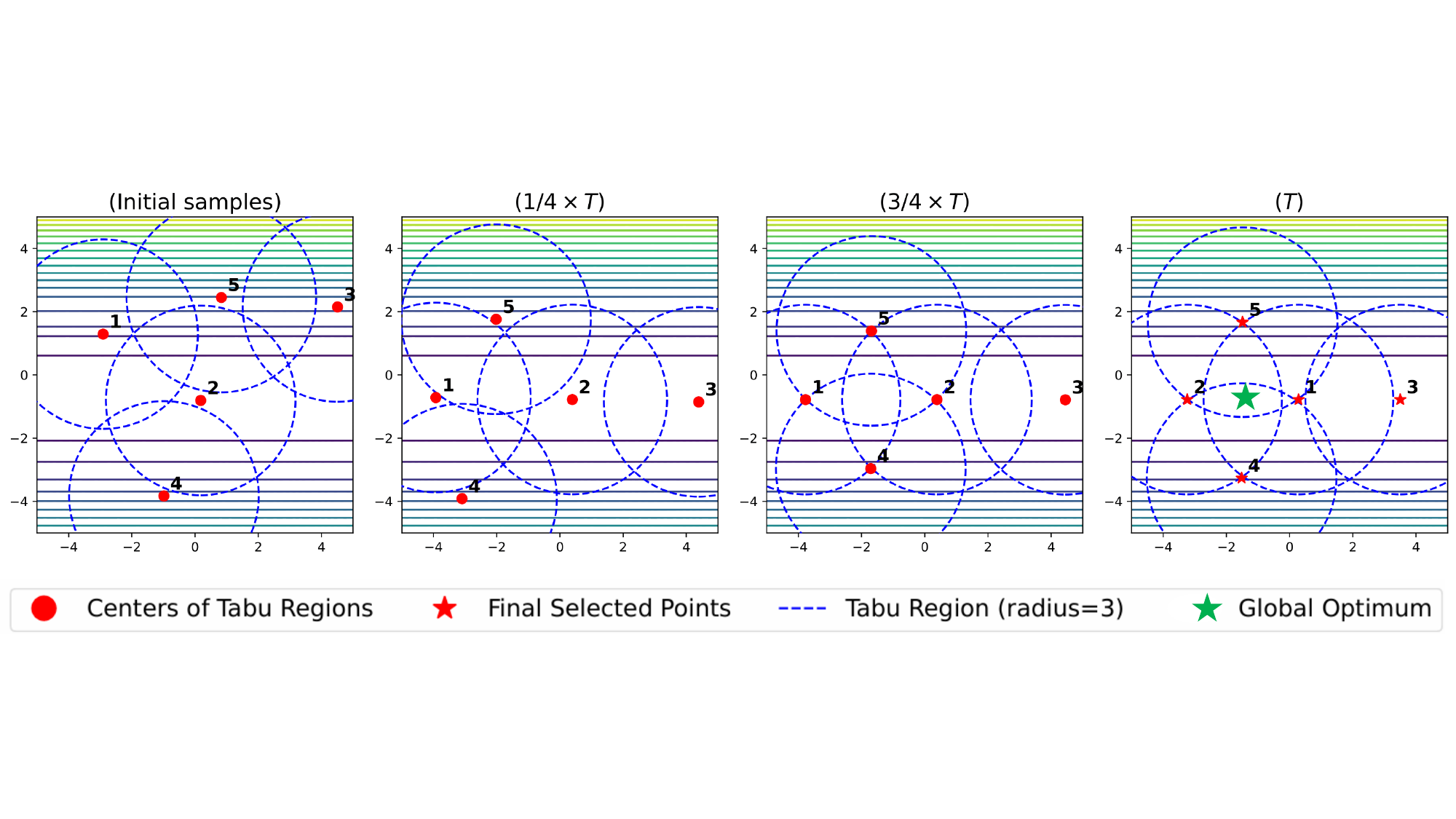}
      \caption{$D = 2$, $T = 1\,000$. Isocontour plots of f2 where dark colors correspond to better values. We show initial means $(x_0)$, tabu region centers after $1/4$ and $3/4$ of the budget, and final selected solutions after clearing. Numbers indicate CMA-ES cascade levels, and circles represent tabu regions. In the last subplot, the green star denotes the true global optimum.}
    \label{fig:algorithm_phases}
\end{figure*}

\newpage
\subsection{Alternative Approaches}
\label{sec:aleternative}

We will compare CMA-ES-DS to the following algorithms:

\textbf{1. Random Sampling}, which samples $T$ points from the search space uniformly at random (i.i.d. sampling).

\textbf{2. Covariance matrix adaptation evolution strategy (CMA-ES)~\citep{hansen_reducing_2003},} evolution strategy that adapts the covariance matrix of a multivariate normal distribution to improve the search for optimal solutions. It iteratively generates new candidate solutions based on the current population and the covariance matrix, updating the covariance matrix and the step size to guide the search towards more promising regions of the solution space. We consider two variants: (i) \textit{single-run CMA-ES}, which performs a single run using the total budget $T$; and (ii) \textit{CMA-ES\_INDEP}, where $k$ independent CMA-ES instances are run in parallel, each with a budget of $T/k$.

\textbf{3. Covariance Matrix Self-Adaptation with Repelling Subpopulations (RS-CMSA)}~\citep{article}, which divides the population into multiple subpopulations that evolve independently, each focusing on a different region of the search space. To maintain diversity and prevent subpopulations from converging toward the same solution, a repelling force is applied whenever two subpopulations get too close to each other. Each subpopulation uses a variant of the CMA-ES algorithm, with self-adapting covariance matrices that allow it to model the local landscape of the problem dynamically. 

\textbf{4. Hill-Valley Evolutionary Algorithm (HillVallEA)}~\citep{maree2019benchmarking}, which
combines the Hill-Valley test~\citep{10.1145/3205455.3205477} with the nearest better tree concept. The Hill-Valley test groups solutions into niches by detecting whether a \textit{hill} exists between them. Intermediate solutions are sampled and clustered using Hill-Valley Clustering, forming niches used to initialize the population for the AMaLGaM-Univariate~\citep{bosman2008matching} core search strategy. The algorithm then refines solutions within each niche using univariate distributions.

\textbf{5. Weighted Gradient and Distance-Based Clustering \newline (WGraD)}~\citep{9002742}. It uses a novel clustering method constructing multiple spanning trees, representing potential niches. The key innovation is the dynamic weighting of gradient and distance factors: for nearby points, the gradient is weighted more heavily, promoting local connectivity, while distant points are weighted more heavily by distance to selectively link them. Then Differential Evolution is used to find the optimum within each niche.

\textbf{6. Accurate Peak Detection in Multimodal Optimization via Approximated Landscape Learning (APDMMO)}~\citep{ma2025accuratepeakdetectionmultimodal}, consisting of three main stages. First, it constructs a surrogate model (Landscape Learner) that uses a neural network with diverse nonlinear activation functions to accurately approximate the objective function. Then, a multi-start AdamW optimizer~\citep{loshchilov2017decoupled} performs fast gradient-based searches on the surrogate to locate potential peaks without extra evaluations. Finally, the detected peaks are refined using parallel SEP-CMA-ES~\citep{ros2008simple} instances to improve solution accuracy.

The first five approaches were also considered in~\citep{santoni2024illuminatingdiversityfitnesstradeoffblackbox}, while APDMMO is a recent MMO method shown to outperform all other baselines on the CEC2013 MMOP benchmark~\citep{li2013benchmark}, with HillVallEA~\citep{maree2019benchmarking} as the next best.

\section{Subset Selection}
\label{sec:3_methods}
We compare a total of three different methods to extract a batch of diverse and high-quality solutions from the full trajectory of all evaluated points; i.e., for the subset selection phase, referred to as Phase~2 in Figure~\ref{fig:process_phases}.  The Gurobi-based method and the greedy algorithm are taken from~\citep{santoni2024illuminatingdiversityfitnesstradeoffblackbox} and are slightly adapted here to our objective of keeping the best-of-all solutions. We also add to our comparison a ``clearing'' approach inspired by~\citep{542703}.
The three methods differ significantly in computational cost and how they balance diversity and solution quality. An empirical comparison of the three methods will be presented in Section ~\ref{sec:compare3method}.

\textbf{Gurobi-based Optimization.} This approach is the computationally most expensive one, both in terms of time and memory. However, it provides an optimal solution to the selection problem, making it the most accurate approach but also the least scalable to higher dimensionalities and different evaluation budgets. 

\textbf{Greedy Selection Approach.} This method, introduced in~\citep{santoni2024illuminatingdiversityfitnesstradeoffblackbox}, greedily swaps points that violate the distance requirements, which are sequentially increased until reaching the $d_{\min}$ limit. Empirical comparison with the Gurobi approach provided in~\citep{santoni2024illuminatingdiversityfitnesstradeoffblackbox} demonstrated this approach to be quite accurate, providing batches that are of not much worse quality than the exact ones extracted using Gurobi. However, the sequential nature of this algorithm implies a computational overhead that is not negligible. 

\textbf{Clearing Process.} This is the fastest and most memory-efficient approach. It works by iteratively picking the point with the lowest fitness value and removing all other points within a specified distance of $d_{\min}$. The process continues until the desired number of points is selected, ensuring that no two points are within the minimum distance from each other. While efficient, it is less likely to find the optimal solution compared to the other methods. 

Regarding the runtime, in our problems, Gurobi runs between almost a minute to several minutes, depending on the problem's dimensionality and budget (e.g., for $D=2$, $T=1\,000$, and $d_{\min} = 1$, runtime $\approx 40$ seconds; for $D = 10$, $T=10\,000$, and $d_{\min} = 10$, runtime $\approx 13$ minutes). The other proposed methods run in the order of seconds and milliseconds. Specifically, for the Clearing method, its runtime is independent of $d_{\min}$ and its range remains in the order of milliseconds depending on dimension $D$ and budget $T$ (e.g., for $D=2$ and $T=1\,000$, runtime $\approx 0.0114$ seconds, while for $D=10$ and $T=10\,000$, runtime $\approx 0.2492$ seconds). On the other hand, the greedy approach's runtime is dependent on $d_{\min}$. It can range from milliseconds to several seconds, or even one minute, for instances with large distances and budgets relative to the dimensionality (e.g., for $D=2$, $T=1\,000$ and $d_{\min} = 1$, runtime $\approx 0.1231$ seconds, while for $D=10$, $T=10\,000$ and $d_{\min} = 10$, runtime $\approx 1$ minute and a few seconds).

\section{Experimental Setup}
\label{sec:exset}
For our experiments, we use the noiseless Black-Box Optimization Benchmarking (BBOB) suite of the COCO environment~\citep{hansen2021coco}.
These functions are defined on the search space $[-5,5]^D$ and are grouped into five categories based on their structural properties: separable functions (f1-f5), functions with low or moderate conditioning (f6-f9), functions with high conditioning and unimodal structure (f10-f14), multimodal functions with clear (f15-f19), and weak global structure (f20-f24).

To keep the computational effort at a reasonable scale, we performed multiple runs on the same instance (instance ID = 0) for each algorithm, with cross-instance robustness of our proposed method ensured by the invariance properties of CMA-ES and further analyzed in Section~\ref{sec:instances}.

\begin{figure*}[t] \center
    \includegraphics[width=\textwidth]{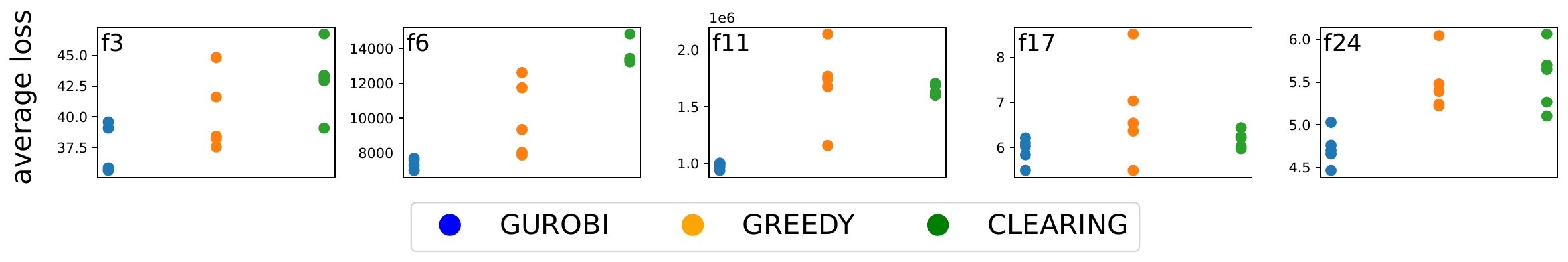}
      \caption{$D = 2$, $T = 10\,000$, $d_{\min} = 3$. Comparison of the subset selection methods, Gurobi-based Optimization, Greedy Approach, and Clearing Process. y-axis represents the average loss of the batch of $k = 5$ points across 5 repetitions (individual dots).}
    \label{fig:compare_3_method}
\end{figure*}

\textbf{Performance Criteria.} 
We measure algorithm performance using \textit{loss}, defined as the absolute difference between the best solution’s quality and the global minimum, which is available for the BBOB functions. 
To align with the goals of this work, we calculate the average loss for the $k$ solutions obtained under a given distance constraint $d_{\min}$. This metric, referred to as \textit{average loss}, is used to assess the quality of the batch of solutions.
In the Appendix, we also report performance based on the individual losses of the batch points, ordering them by quality and then averaging across the five runs.

\textbf{Algorithm Setup.} For the CMAES-DS algorithm, we use the default CMA-ES settings for both the population size $\lambda$ and the stopping criterion. The population size at each generation is set to $\lambda = 4 + \lfloor 3 \log(D) \rfloor$. The size of the \textit{parent} population, i.e., the subset of selected points used to generate the next generation in each iteration, is computed as $\mu = \lfloor \lambda / 2 \rfloor$. The stopping criteria include several default conditions, such as \texttt{TolX} $= 10^{-11}$, \texttt{TolFun} $= 10^{-11}$, \texttt{TolFunHist} $= 10^{-12}$, \texttt{TolFunRel} = 0, \texttt{TolStagnation} = 146 iterations and \texttt{MaxIter} $= 10^3 \cdot D^2$. The initial step-size in each coordinate is set to 1 based on recommendations from~\citep{hansen_reducing_2003}.
For all other algorithms, we use their default hyperparameters as provided by their respective implementations. Each algorithm is executed independently for 5 runs.

\textbf{Evaluation Setup.} To evaluate the potential of each algorithm under distance constraints, we apply the clearing method to their respective histories. This ensures that the solutions selected for the final batch meet the minimum pairwise distance constraint $d_{\min}$.
Experiments are conducted across three search space dimensions, $D = 2, 5, 10$. Two total evaluation budgets are considered: $T = 1\,000$, and $T = 10\,000$ function evaluations.
Various $d_{\min}$ are explored: for $D = 2 : d_{\min} \in \{1, 3, 5\}$, for $D = 5$ and $D = 10 : d_{\min} \in \{1, 3, 5, 6, 10, 12\} $.
Note that large values of $d_{\min}$, such as 10 or 12, approach the scale of the search space $[-5,5]^D$, which can significantly constrain the search space and impact performance (especially at $D = 5$).
Because APDMMO is very expensive to run (about 30 h in CPU time for one BBOB function) and performed poorly for our scope at $T = 1\,000$, we decided not to run it at $T = 10\,000$.

We report only the results that support the main observations; additional plots are provided in the Appendix, and the complete set is available on GitHub~\citep{code}.

\section{Results and Discussion}
\label{sec:results}

\subsection{Comparing Three Subset Selection Approaches}

To compare the three subset selection methods (Gurobi, Greedy, and Clearing), we applied them to the portfolios obtained from the optimization histories of five independent CMA-ES-DS runs in 2D, with parameters $T=10\,000$, $k=5$, and $d_{\min}=3$. We limited the analysis to $D=2$ due to the high computational cost of Gurobi in higher dimensions.

Figure ~\ref{fig:compare_3_method} shows the average loss for the three methods across five repetitions. Since the results are homogeneous across functions belonging to the same BBOB category, we only show results for one representative function per BBOB group. Generally, Gurobi achieves the lowest average loss, as it guarantees the optimal solution to such combinatorial selection problems, unlike heuristic methods like Greedy and Clearing, which do not guarantee global optimality. Second ranked is the Greedy approach, while Clearing performs worst but remaining computationally cheapest. Interestingly, for some functions such as f11 and f17, Clearing outperforms Greedy on average, indicating that simpler methods can occasionally yield better batches depending on the function landscape. 
We also observe that the choice of selection method influences the results more than the specific CMA-ES-DS runs. Gurobi yields the most consistent outcomes with minimal variance, while Greedy shows the highest variability and Clearing is intermediate. This highlights that selection strategy strongly impacts the stability of the final solutions.
Overall, although Gurobi delivers the best solution quality, its computational cost is prohibitive. Since the performance of Greedy and Clearing generally falls within a similar numerical range, we proceed with Clearing for the remainder of this study, as it is more stable and computationally efficient than the alternative option.

\label{sec:compare3method}

\subsection{Comparing Portfolios Generated by the Algorithms}
\label{sec:compareportfolio}
\subsubsection{Summary View}
\begin{figure}[t] \center
    \includegraphics[width=\textwidth]{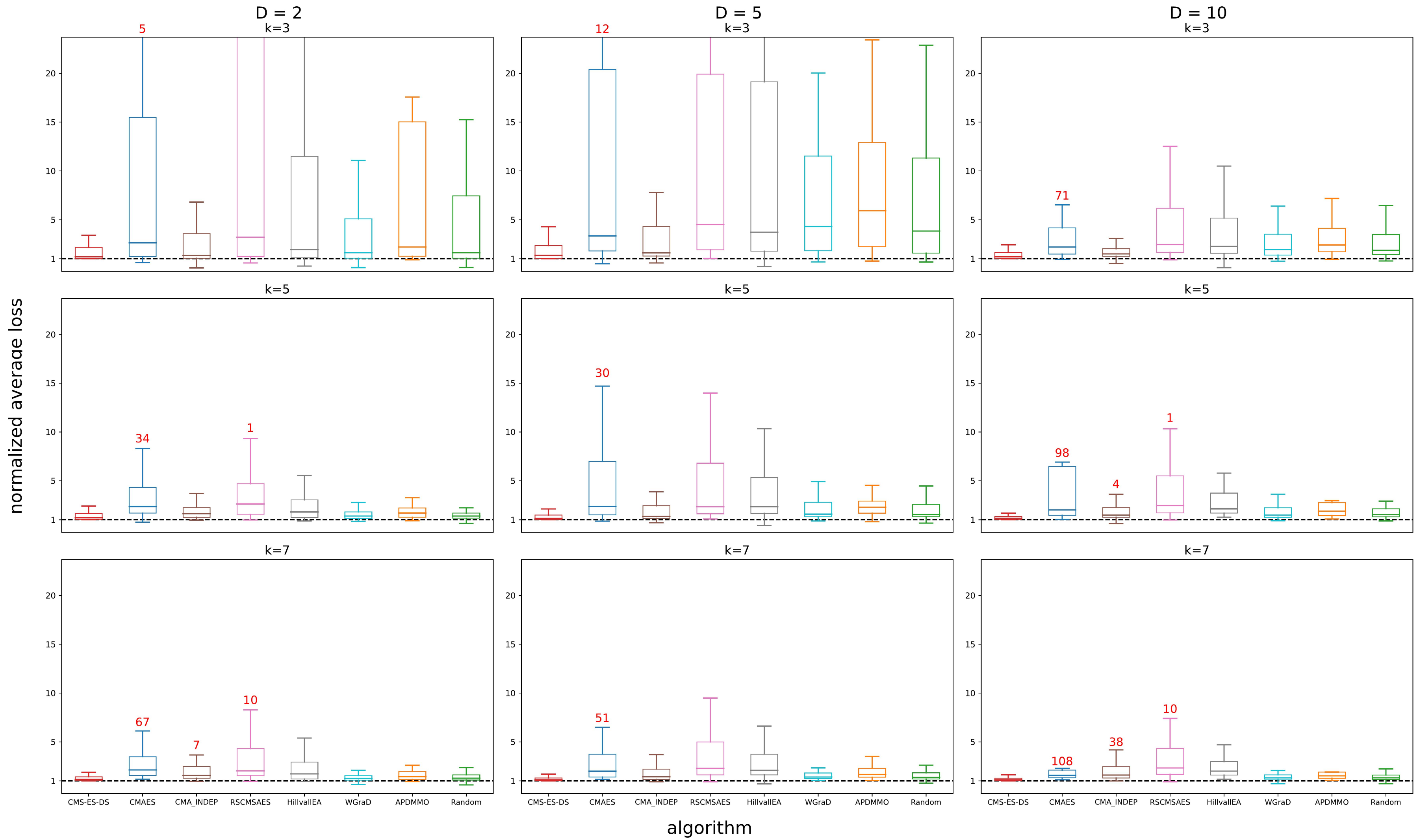}
     \caption{
$T = 1\,000$. Normalized average loss of the final batch of $k$ points ($k = 3, 5, 7$) aggregated over the 24 BBOB functions and 5 repetitions for each algorithm. Results are shown for three different settings: $D = 2$, $d_{\min} = 3$; $D = 5$, $d_{\min} = 5$; and $D = 10$, $d_{\min} = 10$. 
For each function, the final batch loss of each algorithm is normalized by the
best CMA-ES-DS value across its 5 runs. Values below 1 (dashed black line) indicate
better performance than CMA-ES-DS.
Outliers are not displayed for clarity. 
For dimension 2, some boxplots are cut to keep a common scale.
Red numbers displayed above the boxplots indicate the failed runs (out of
120), resulting in statistics over less runs.}

\label{fig:boxplotall}
\end{figure}
Figure~\ref{fig:boxplotall} provides an overview of the normalized average loss achieved by each algorithm for the final batch of $k$ points ($k = 3, 5, 7$) across three problem dimensionalities: $D = 2$, $D = 5$, and $D = 10$. The minimum required distance is set proportionally to dimensionality: $d_{\min} = 3$, $5$, and $10$, for dimension $2$, $5$, and $10$, respectively. We analyze the behavior under a fixed budget of $T = 1\,000$ function evaluations. Each boxplot aggregates results over the 24 BBOB benchmark functions and 5 independent runs (120 data points per algorithm when all runs complete successfully), normalized by the
best CMA-ES-DS value across its 5 runs.
For each function, all values are divided by the best final batch loss achieved by CMA-ES-DS over its 5 runs, so that values below $1$ indicate better average performance compared to CMA-ES-DS.
The red numbers above the boxplots represent the number of failed runs, i.e., cases in which an algorithm could not generate a full batch of $k$ diverse solutions satisfying the minimum distance constraint.

Overall, CMA-ES-DS exhibits the most stable behavior, as indicated by the shortest
interquartile ranges (IQRs) across all dimensions. 
In $D=2$, we observe a noticeably larger variance for all algorithms, 
likely due to the stronger impact of the distance constraint 
($d_{\min}=3$) when placing several well-performing points in a low-dimensional space. 
Also in this setting, our algorithm maintains a small variance.
In general, the medians and most of the IQRs of the competing algorithms remain
above the dashed black line ($y=1$), suggesting the advantage of CMA-ES-DS, though the lower whiskers of some algorithms cross below this line.
Such cases are investigated in more detail in the following sections.

For all dimensionalities, as the batch size increases, we observe a higher number of failures for single-run CMA-ES, CMA-ES\_INDEP, and RS-CMSA. This can be attributed to the strong tendency of these algorithms to concentrate the search locally around a single global optimum. 
Indeed, despite being an MMO algorithm, RS-CMAES applies a CMA-ES variant within each detected basin of attraction, potentially overlooking multiple high-quality solutions within the same basin. Notably, for CMA-ES\_INDEP and RS-CMSA, no failures are observed for $k = 3$, indicating that the algorithms’ histories include at least three points that meet the minimum distance requirement.

Also, we observe that single-run CMA-ES and RS-CMSA generally exhibit the highest variance among all methods, indicating high variability across runs.

\begin{figure}[t] \center
    \includegraphics[width=\textwidth]{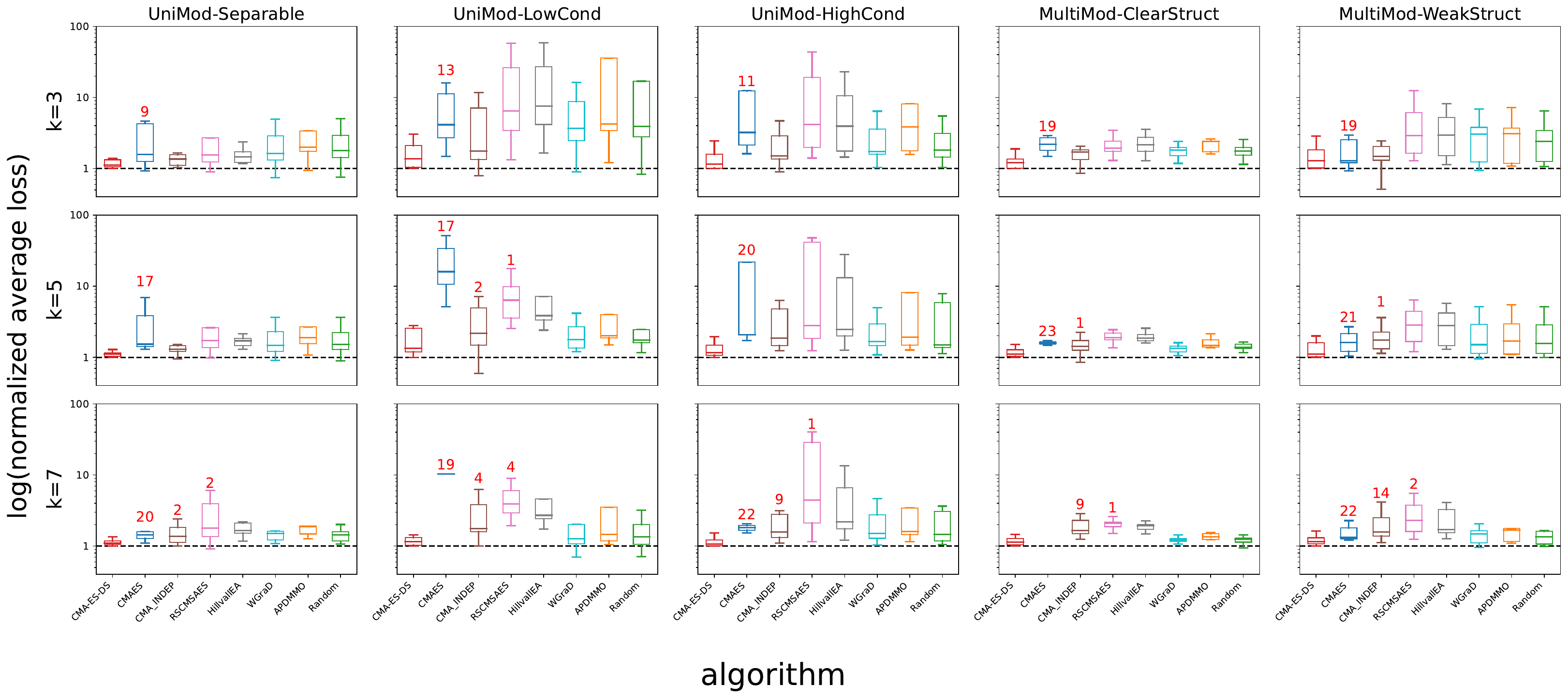}
      \caption{$D = 10$, $T = 1\,000$, $d_{\min} = 10$. Normalized average loss of the final batch of $k$ points ($k = 3$, $k = 5$, $k = 7$), aggregated over the five groups of BBOB functions and 5 repetitions for each algorithm. The y-axis is shown on a logarithmic scale. For each function, the final batch loss of each algorithm is normalized by the
best CMA-ES-DS value across its 5 runs. Values below 1 (dashed black line) indicate better performance than CMA-ES-DS. Outliers are not displayed for clarity. Red numbers displayed above the boxplots indicate the failed runs (out of 25 for all the groups but UniMod-LowCond, which is out of 20), resulting in statistics over less runs.}
    \label{fig:boxplotsgroups}
\end{figure}

Figure~\ref{fig:boxplotsgroups} shows a distributed version of this analysis, organized by function groups focusing on the setting $D = 10$, $d_{\min} = 10$.
We observe that MMO algorithms generally perform better on multimodal functions, particularly those with a clear global structure, where they also show a small variance. Their effectiveness slightly drops on functions with weak global structure, being the landscapes more difficult to navigate.

In contrast, MMO algorithms perform worse on the first three groups---separable, low-conditioning, and high-conditioning functions--- which are mainly composed by unimodal landscapes. The bad performance compared to our method is especially evident for functions with large basins of attraction and moderate or high conditioning (2nd and 3rd columns). Here, MMO algorithms typically focus on finding the global optimum within the basin of attraction, while minimizing exploration of suboptimal areas within the same basin. This is evident with RS-CMSA, which employs CMA-ES within each basin. Due to CMA-ES’s limited exploration, enforcing a minimum pairwise distance smaller than the basin size can lead to suboptimal final batches: the search trajectory does not provide enough diverse candidates within the basin, and the batch selection procedure must select points from sparser, less promising regions outside the distance constraint.
Simpler methods like Random Search or WGraD often perform better. WGraD, leveraging weighted gradients and clustering via spanning trees, has shown strong exploratory capabilities in high dimensions~\citep{9002742}, outperforming classical niching methods.

It is also worth noting that, for $k = 3$, CMA-ES\_INDEP occasionally performs below the baseline ($y=1$) in some function groups. This suggests that independently running multiple CMA-ES instances can outperform a single-run in finding good solutions when a level of diversity is required among them. Even though it may localize the initial $k$ candidates in similar regions, the use of multiple independent CMA-ES instances introduces more diversity than a single CMA-ES run. However, its performance tends to degrade as $k$ increases and the task requires more extensive exploration of the search space. In such cases, applying the cascading mechanism, as employed in CMA-ES-DS, becomes more effective by promoting more diverse sampling rather than relying solely on $k$ independent CMA-ES instances.

Single-run CMA-ES fails to extract full batches across all settings, which is consistent with its focus on a single optimum. Notably, for $k=7$ and multimodal functions with clear structure, it fails in all runs (no box plot shown), underlining the difficulty in satisfying the constraint. For this algorithm, low IQRs are misleading, as this is due to the few successful runs. CMA-ES\_INDEP fails less frequently then single-run CMA-ES and only starting from $k=5$, confirming that using $k$ independent instances increases diversity and helps meet the diversity constraint more effectively. RS-CMSA performs slightly better, typically succeeding at $k=3$, but starts failing at $k=5$ (especially in low-conditioning cases), with failures widespread at $k=7$.

CMA-ES-DS consistently performs well across all groups and batch sizes, outperforming all other algorithms and demonstrating robustness and strong adaptability to diverse landscapes.

\subsubsection{Closer Examination}
\begin{figure}[t] \center
    \includegraphics[width=\textwidth]{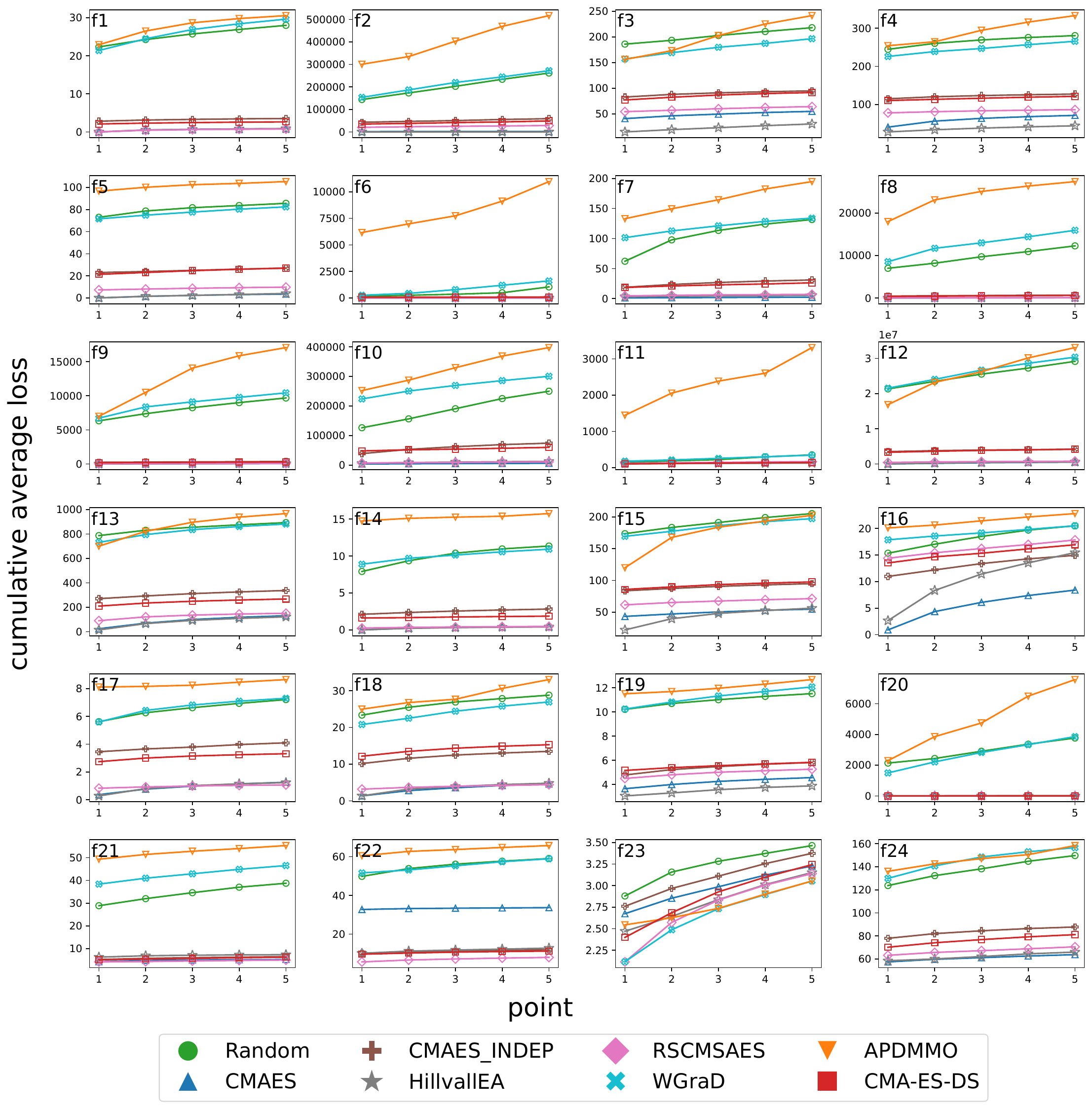}
      \caption{$D = 10$, $T = 1\,000$, $d_{\min} = 1$. Cumulative average loss across the $k = 5$ points in the batch for the 24 BBOB functions. Each subplot corresponds to a specific BBOB function, and the x-axis represents the batch points ($x=1$ to $x=5$). The y-axis shows the cumulative average loss, where each curve represents the mean performance of an algorithm over 5 independent runs.}
    \label{fig:D101Kd1}
\end{figure}

\begin{figure}[t] \center
    \includegraphics[width=\textwidth]{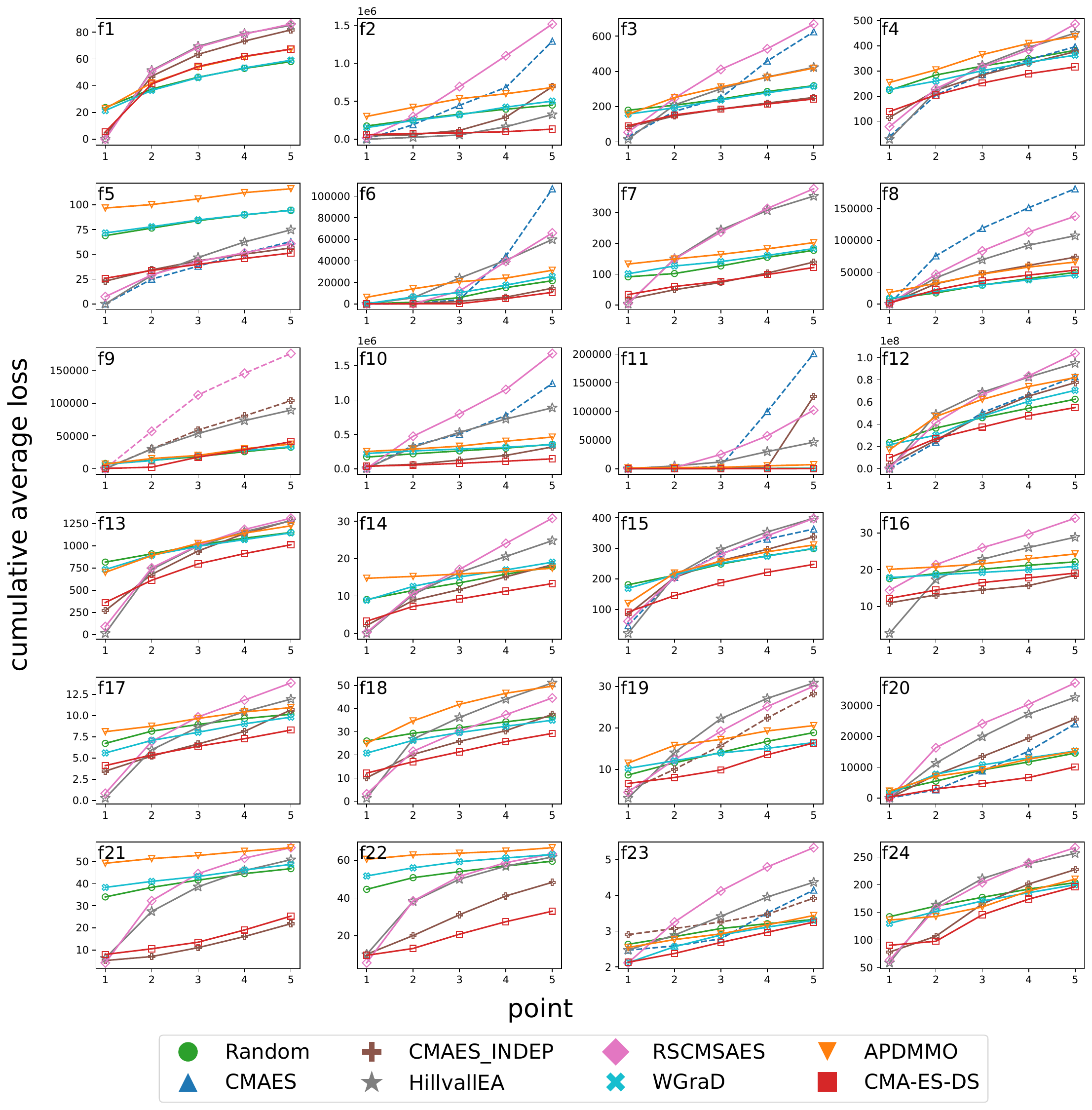}
      \caption{$D = 10$, $T = 1\,000$, $d_{\min} = 10$. Same as Figure ~\ref{fig:D101Kd1}, but for $d_{\min} = 10$.}
    \label{fig:D101Kd10}
\end{figure}

Results across the experimental settings demonstrate that our proposed approach, CMA-ES-DS, performs well, achieving competitive or superior performance compared to the other algorithms. This holds across all the various dimensions, distance constraints, and evaluation budgets tested, with one exception. Specifically, for high-dimensional search spaces ($D = 10$), small budgets ($T = 1\,000$), and minimal distance constraints ($d_{\min} = 1$), alternative algorithms such as basic single-run CMA-ES, RS-CMSA, and HillVallEA outperform CMA-ES-DS (Figure ~\ref{fig:D101Kd1}). 
This outcome is expected, as these algorithms are not designed to enforce distance constraints and thus excel in locating high-quality solutions in global optimization tasks without strong spatial diversification.
This pattern aligns with findings in~\citep{santoni2024illuminatingdiversityfitnesstradeoffblackbox}. For instance, similar behavior would likely be observed for $d_{\min} = 0.5$ in $D = 2$ or $D = 5$. However, as the distance constraint becomes more stringent, our approach outperforms others due to its explicit handling of diversification. This trend is particularly evident in Figure ~\ref{fig:D101Kd10}, where increasing $d_{\min}$ to 10 leads to better results for CMA-ES-DS compared to all the other algorithms.

\begin{figure}[t] \center
    \includegraphics[width=\textwidth]{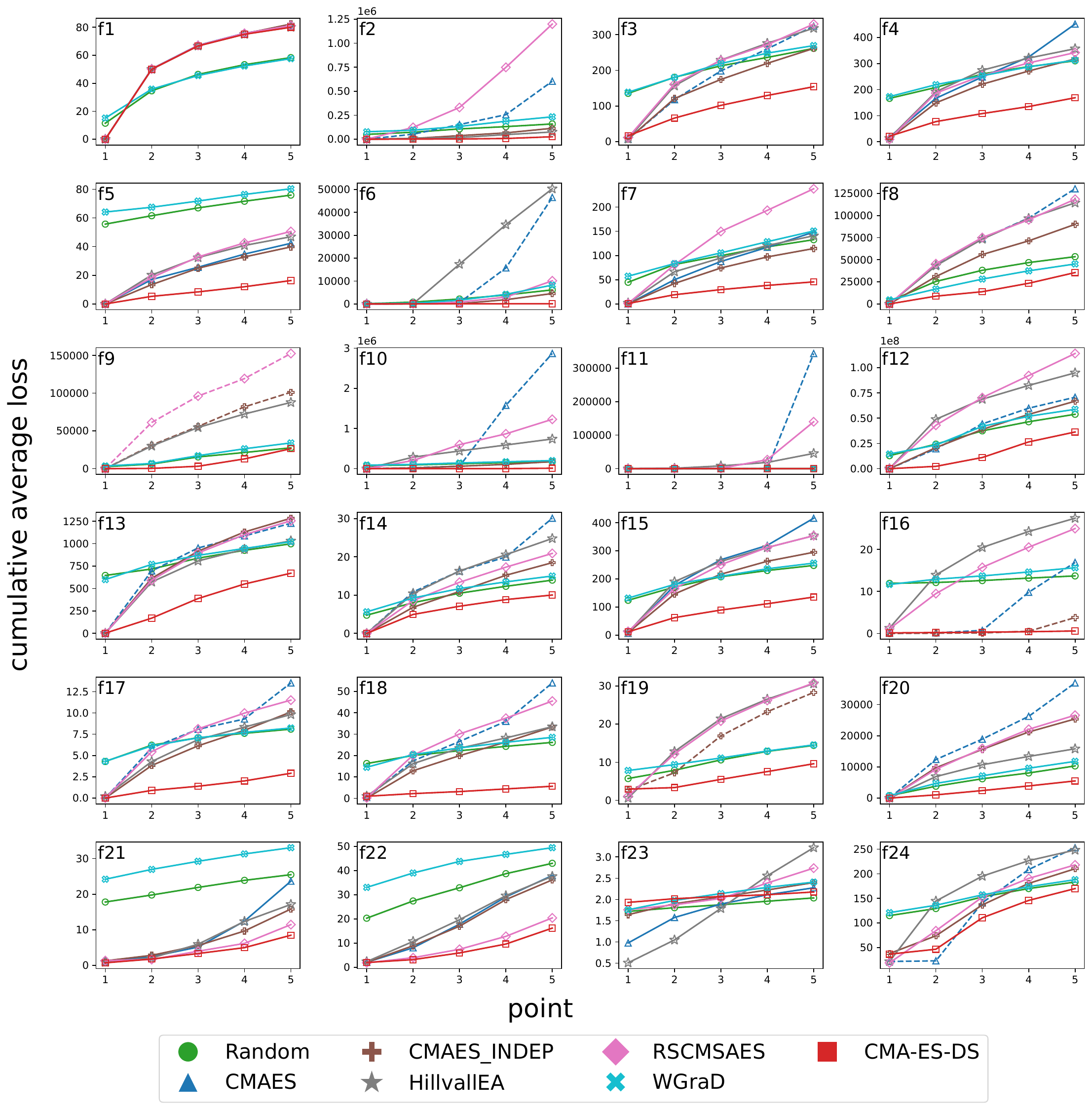}
      \caption{$D = 10$, $T = 10\,000$, $d_{\min} = 10$. Same as Figure ~\ref{fig:D101Kd10}, but for initial portfolio size $T = 10\,000$.}
    \label{fig:D1010Kd10}
\end{figure}

An additional observation deducted from Figure ~\ref{fig:D101Kd10} is the performance dependency on the budget $T$. For a small budget ($T = 1\,000$), the first solution identified by CMA-ES-DS is often suboptimal because the approach divides the budget equally across the diverse solutions, allocating scarce resources to the various CMA-ES instances. This contrasts with CMA-ES and MMO algorithms, which focus solely, on finding one or multiple optima respectively, without imposing distance constraints. Instead, for a larger budget $T = 10\,000$ CMA-ES-DS demonstrates good results also in the quality of the leader point while maintaining its advantage in providing diversified alternatives (Figure ~\ref{fig:D1010Kd10}).

Reliability is another strength of CMA-ES-DS, as evidenced by Figure ~\ref{fig:D101Kd10} and~\ref{fig:D1010Kd10}. In scenarios with strong distance constraints ($d_{\min} = 10$), some algorithms fail to identify a set of $k = 5$ solutions at the required distances across their histories. This failure is illustrated by dashed (some failed runs) or missing lines (all failed runs). Notably, algorithms based on CMA-ES and RS-CMSA often suffer from this limitation, as they tend to localize the search without adequate exploration. Their exploitative nature would leave areas of the search space, where diverse solutions could be located, completely unexplored. In contrast, CMA-ES-DS consistently identifies feasible solutions, demonstrating its reliability. For example, in Figure ~\ref{fig:D101Kd10} we have a missing or a dashed line for all functions of CMA-ES and f9 of RS-CMSA. A similar behavior is shown in Figure ~\ref{fig:D1010Kd10}, with fewer failures for CMA-ES due to the larger available budget.

Figure ~\ref{fig:D101Kd10} and~\ref{fig:D1010Kd10} also confirm the observation from~\citep{santoni2024illuminatingdiversityfitnesstradeoffblackbox} that random sampling outperforms MMO algorithms under stronger distance constraints; because the attraction basins identified by the MMO algorithms are at distances that are smaller than those enforced by the distance constraint.  

Although our approach consistently outperforms random sampling, the relative advantage decreases as $d_{\min}$ increases. For larger distance constraints, algorithms achieving a better coverage of the search space, such as random sampling, tend to perform well due to their inherent exploratory nature. 

Moreover, CMA-ES-DS is particularly beneficial when searching for diverse solutions in settings where the ratio between dimensions and total budget is large. Otherwise, all methods would achieve a sufficient coverage of the search space and perform similarly, reducing the advantage of using more sophisticated algorithms. In addition, when this ratio is large, the advantages of our approach are more pronounced for larger $d_{\min}$, where our explicit diversification strategy outperforms methods that focus solely on global optimization. As shown by our results, the benefits of CMA-ES-DS are most pronounced for $D = 5$ (see Figure ~\ref{fig:D51Kd5} in the Appendix) and $D = 10$ (see Figure ~\ref{fig:D101Kd10} and~\ref{fig:D1010Kd10}). For $D = 2$, mainly with $T = 10\,000$, all methods perform comparably (see Figure ~\ref{fig:D210Kd3} in the Appendix).
However, we can observe some exceptions for two specific landscapes. For unimodal and low-conditioned landscapes such as the sphere function (f1), random sampling can outperform our method even where this ratio is large (see Figure ~\ref{fig:D101Kd10} and~\ref{fig:D1010Kd10}). This occurs because random sampling tends to find a best solution of lower quality compared to other optimizers, which allows the remaining solutions to be placed on lower isocontour levels. In fact, we observe in both Figure ~\ref{fig:D101Kd10} and Figure ~\ref{fig:D1010Kd10} that on f1 random sampling achieves a worse global solution than the other methods, but batches of better average quality. However, this phenomenon is observed only when the distance constraint is significantly strong. For small distances (e.g., Figure ~\ref{fig:D101Kd1}), CMA-ES-DS significantly outperforms random sampling.
This suggests that relaxing the requirement to select as $x^1$ the best evaluated solution could improve its performance for such cases when distance constraints are strong.

In repetitive landscapes like f23, the performance of CMA-ES-DS varies significantly depending on the budget. For $D = 10$, $T = 1\,000$ and $d_{\min} = 10$ (Figure ~\ref{fig:D101Kd10}), CMA-ES-DS confirms to be the best-performing method, given the large ratio between dimensionality and total budget. For $T = 10\,000$ (Figure ~\ref{fig:D1010Kd10}), random sampling outperforms CMA-ES-DS, as the high budget allows random sampling to adequately cover the landscape, which is particularly beneficial in repetitive landscapes where the risk to converge to local optima is higher. In contrast, for $T = 1\,000$ (Figure ~\ref{fig:D101Kd10}), random sampling fails to achieve comparable performance due to the limited number of function evaluations. In this case, CMA-ES-DS stands out as it uses efficiently the small budget to identify diverse and high-quality solutions under distance constraints.
Moreover, for $T = 10\,000$ (Figure ~\ref{fig:D1010Kd10}), basic CMA-ES and HillvallEA identify a very good leader solution, highlighting their global optimization strengths. On the other hand, CMA-ES-DS produces flatter curves, reflecting the fact that it is able to detect various basins of attraction of the multimodal landscape. Notably, for $T = 1\,000$ and $d_{\min} = 10$, CMA-ES-DS emerges as the best-performing method, as it effectively balances the constraints of limited evaluations and the requirement for diverse solutions.

For the most favorable setup---high dimension $D = 10$, small budget $T = 1\,000$, and high $d_{\min} = 10$---we also conducted a robustness analysis (see Figure ~\ref{fig:variance} in the Appendix). Our method demonstrates consistent performance across different functions, avoiding the large fluctuations observed in other algorithms, which may show stable results on some functions but significantly greater dispersion on others. From this figure, we also observe that, in low-budget settings, our method does not always outperform the baselines in terms of the leader solution’s performance. However, it consistently achieves the lowest average loss across the final batch.

\begin{figure}[t] \center
\includegraphics[width=\textwidth]{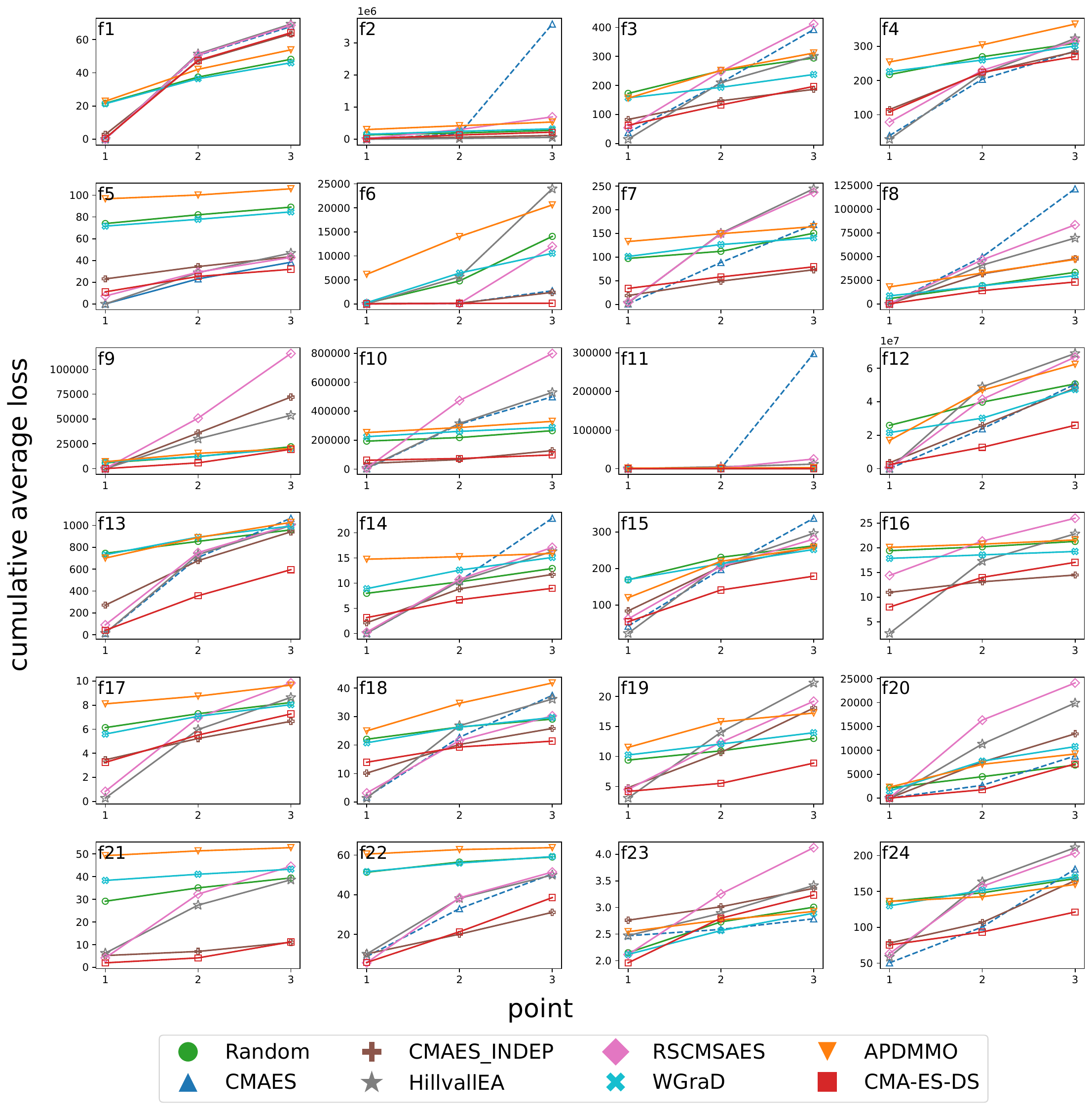}
      \caption{$D = 10$, $T = 1\,000$, $d_{\min} = 10$, $k = 3$. Cumulative average loss across the $k = 3$ points in the batch for the 24 BBOB functions. Each subplot corresponds to a specific BBOB function, and the x-axis represents the batch points ($x=1$ to $x=3$). The y-axis shows the cumulative average loss, where each curve represents the mean performance of an algorithm over 5 independent runs.}
    \label{fig:D101Kd10k3}
\end{figure}

\begin{figure}[t] \center
    \includegraphics[width=\textwidth]{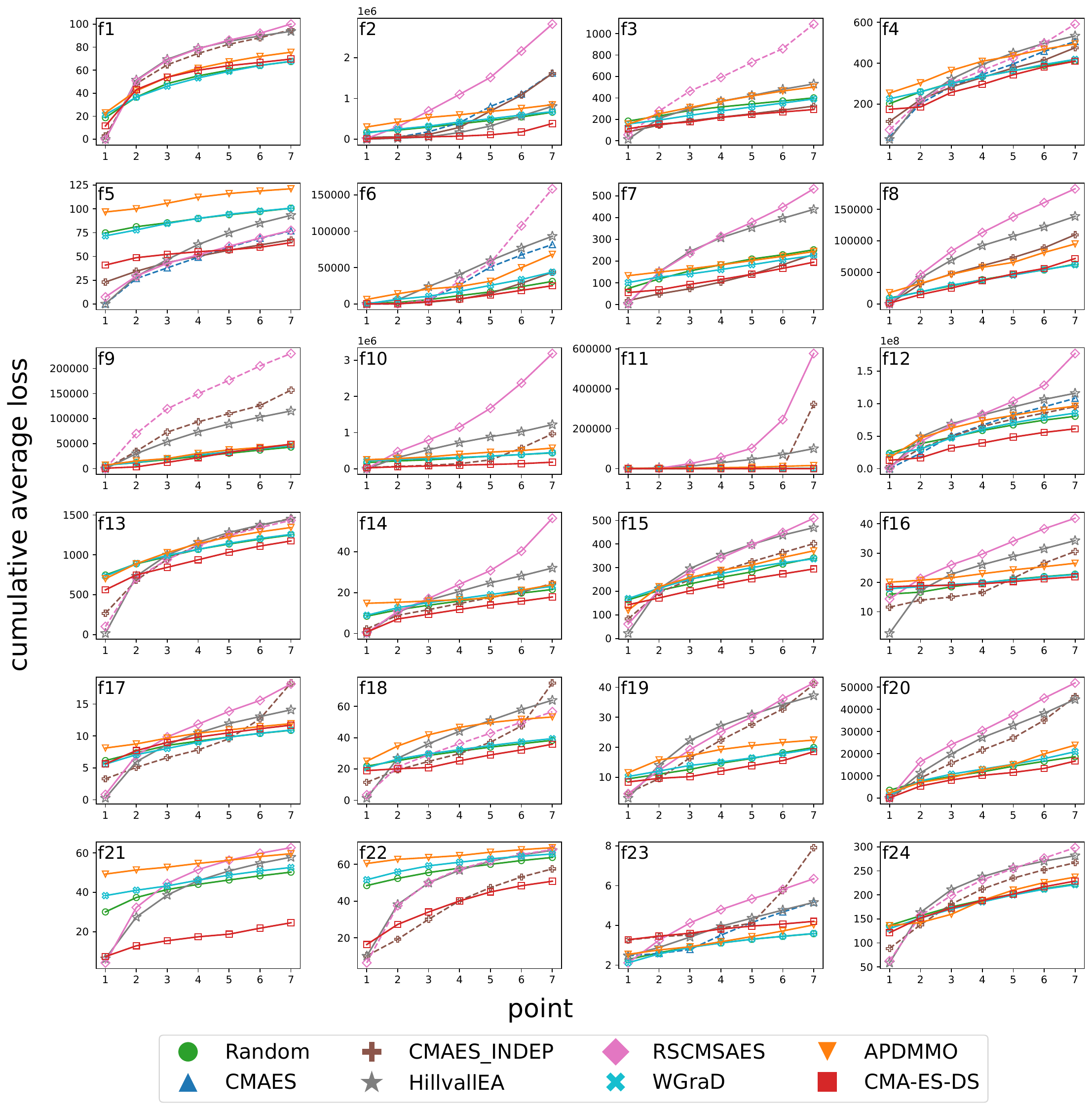}
      \caption{$D = 10$, $T = 1\,000$, $d_{\min} = 10$, $k = 7$. Cumulative average loss across the $k = 7$ points in the batch for the 24 BBOB functions. Each subplot corresponds to a specific BBOB function, and the x-axis represents the batch points ($x=1$ to $x=7$). The y-axis shows the cumulative average loss, where each curve represents the mean performance of an algorithm over 5 independent runs.}
    \label{fig:D101Kd10k7}
\end{figure}

\subsection{Impact of different Batch Size \texorpdfstring{$k$}{k}}

In this section, we investigate the performance of our method with different batch sizes, specifically $k = 3$ and $k = 7$, using the previously identified favorable setting: $D = 10$, budget $T = 1\,000$, and $d_{\min} = 10$ (Figure ~\ref{fig:D101Kd10}). Figures~\ref{fig:D101Kd10k3} and~\ref{fig:D101Kd10k7} show the results for $k = 3$ and $k = 7$, respectively.
Overall, we observe that our algorithm consistently outperforms or is comparable to the baselines across different values of $k$, with the algorithm ranking remaining largely stable, with only a few exceptions.

\textbf{Batches of 3 points.} For $k = 3$ (Figure ~\ref{fig:D101Kd10k3}), the performance curve contains only the initial part compared to what we observed in Figure ~\ref{fig:D101Kd10}, as only three points are evaluated. As previously discussed, the first solution identified by CMA-ES-DS is often inferior compared to those of CMA-ES and MMO algorithms, indicating weaker exploitation capabilities. Consistently, the advantage of our method is less pronounced for $k = 3$ compared to $k = 5$. 

For example, in f4, f6, and f20, CMA-ES and HillVallEA, perform comparably or better than CMA-ES-DS in leader quality, while in the same benchmarks our method shows clear advantage with larger $k$ (Figure ~\ref{fig:D101Kd10}).

Notably, CMA-ES appears more competitive at $k = 3$ because it fails less often to meet the distance constraint: in Figure ~\ref{fig:D101Kd10k3} its performance is often visible, though partially dashed, indicating some failed runs, unlike in Figure ~\ref{fig:D101Kd10}, where its curve is often absent due to consistent failure.

Despite this, our method generally maintains superior performance and achieves even greater margins on functions f8, f9, f12, and f13 compared to the $k = 5$ case.

\textbf{Batches of 7 points.} For $k = 7$, the performance gap between CMA-ES-DS and the other baselines tends to decrease across all the functions, compared to $k = 5$. 

This behavior can be attributed to the cascading approach on which our algorithm is based. As we request more points, additional constraints progressively restrict the feasible region for the cascading CMA-ES instances.

\subsection{Impact of different budget \texorpdfstring{$T$}{T} }
We analyze the evolution of the CMA-ES-DS's performance with respect to the budget~$T$, specifically $T \in \{1\,000,\ 3\,000,\ 5\,000,\ 10\,000\}$, fixing the dimension to 10, the distance constraint to $d_{\min} = 10$, and the batch size to $k = 5$.
\begin{figure}[t] \center
    \includegraphics[width=\textwidth]{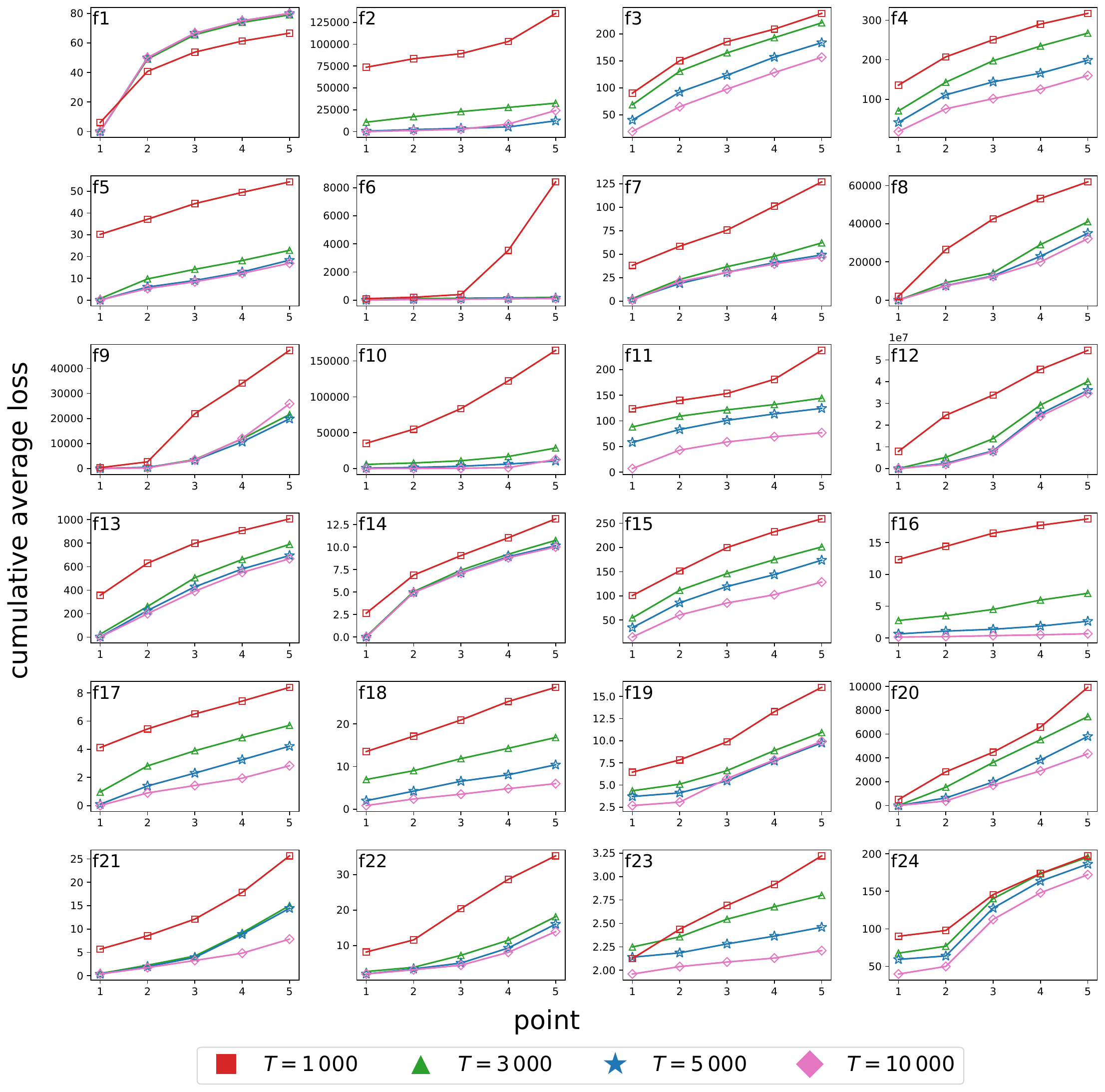}
     \caption{$D = 10$, $d_{\min} = 10$, $k = 5$. Cumulative average loss across the $k = 5$ points in the batch for the 24 BBOB functions, for different portfolio sizes $T \in \{1\,000,\ 3\,000,\ 5\,000,\ 10\,000\}$. Each subplot corresponds to a specific BBOB function, and the x-axis represents the batch points ($x=1$ to $x=5$). The y-axis shows the cumulative average loss, where each curve represents the mean performance over 5 independent runs of CMA-ES-DS.}

    \label{fig:intermediateT}
\end{figure}

Figure~\ref{fig:intermediateT} shows that the quality of the selected batch—both in terms of the leader and the full set of $k = 5$ points—generally improves with increasing budget, and so portfolio size~$T$. This is expected, as a larger initial portfolio allows for a more thorough exploration of the search space. An exception is observed on function f1, where the performance degrades increasing~$T$ and the quality of the leader due to the difficulty in placing the points on good isocontour level when having a optimal leader as previously discussed.

Moreover, the performance gap between $T = 1\,000$ and $T = 5\,000$ is substantially larger than the one between $T = 5\,000$ and $T = 10\,000$, which in some cases is almost negligible (e.g., f2, f5, f12, f14). This suggests that increasing the budget beyond a certain point yields diminishing returns. Since the space has already been sufficiently explored at $T = 5\,000$, further increasing $T$ results in only marginal gains while consuming additional evaluations and CPU time—resources that could be conserved.

The intermediate setting $T = 3\,000$ sometimes offers a performance in between the two adjacent budgets (e.g., f3, f4, f20, f23, f24), while in others (e.g., f2, f9, f10, f12, f22) it performs nearly as well as $T = 5\,000$. This highlights the potential of using intermediate portfolio sizes to save computational budget depending on the problem landscape.

\subsection{CMA-ES-DS on different instances}
\label{sec:instances}

\begin{figure}[t] \center
    \includegraphics[width=\textwidth]{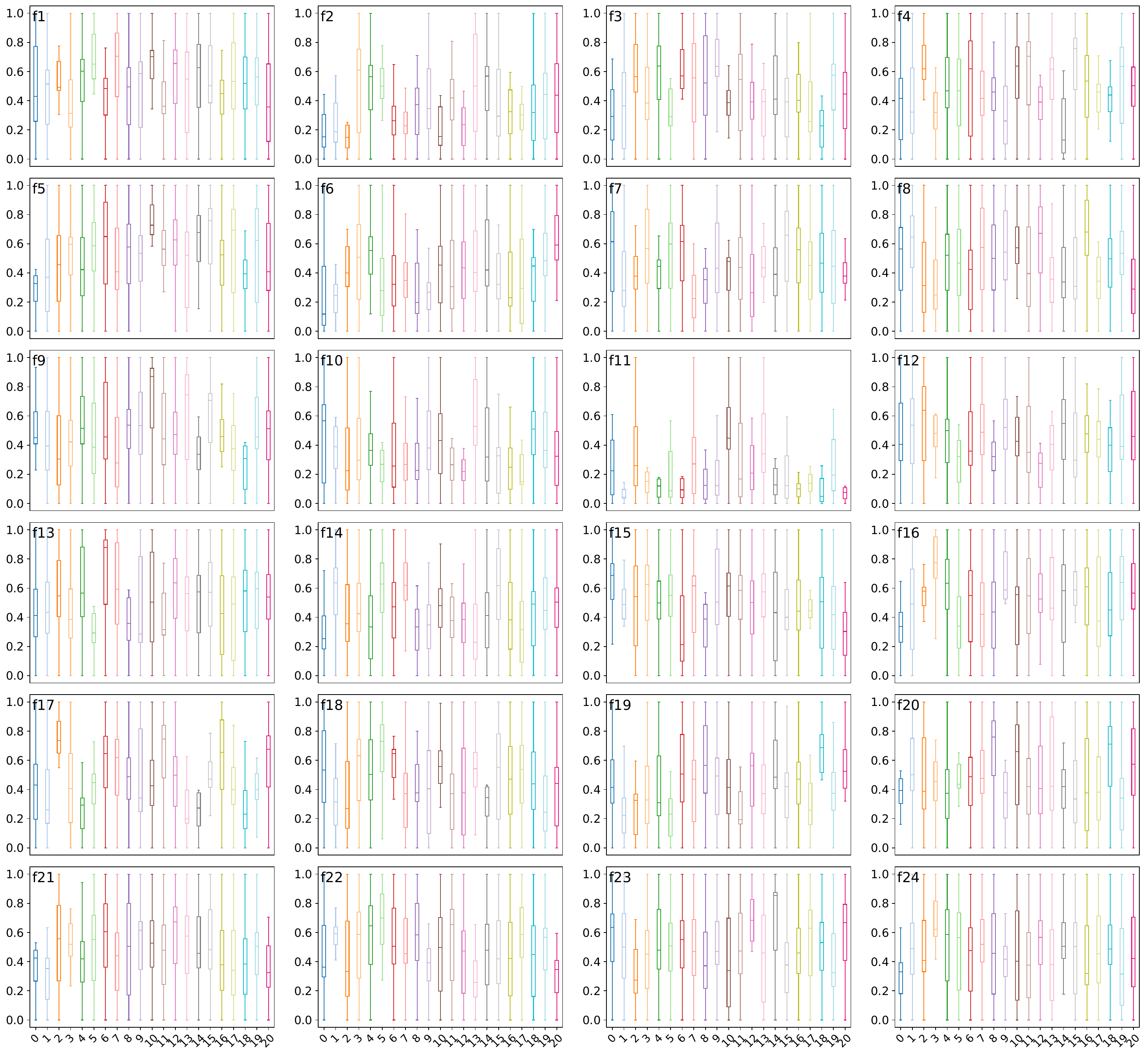}
      \caption{Boxplots of the normalized performance values obtained by CMA-ES-DS on 21 different instances of each of the 24 BBOB problems. Each subplot corresponds to one problem; the x-axis shows the instance number (from 0 to 20), and each colored box represents the distribution across fifteen independent runs. The y-axis shows the average performance of the final batch, normalized using min–max scaling independently for each problem and instance.
}
    \label{fig:instances}
\end{figure}

To discuss the robustness of the performance of CMA-ES-DS, we tested it on all 24 BBOB problems across 21 different instances each. These instances are generated through transformations such as shifts, rotations, and rescalings applied to the base instance (instance 0), which is consistently used throughout this paper for reference. Figure~\ref{fig:instances} shows boxplots of the results for each problem, where the y-axis reports the normalized performance using min-max scaling (applied per problem and instance), and the x-axis lists the 21 instances. Each box summarizes fifteen independent runs for that specific instance.

Figure~\ref{fig:instances} shows that, while some variability exists across instances, the performance of CMA-ES-DS on instance 0 is generally aligned with the rest. Specifically, instance 0 is neither systematically better nor worse than other instances, and its performance lies close to the median in most cases. This supports the idea that results obtained on instance 0 are representative of the algorithm's behavior across transformed instances of the same problem. 
The variability observed, including that of instance 0, is largely due to the stochastic nature of CMA-ES and the differences in initialization across multiple runs. This intrinsic variance is expected and consistent with the algorithm's design. These observations validate our use of instance 0 as a proxy in the paper.

To complement the boxplots and further assess the robustness of the algorithm across different problem instances, we performed paired \textit{t}-tests comparing the performance on instance 0 with each of the other 20 instances, for each problem separately. 
We used a significance level of $\alpha = 0.05$ and computed the p-values using a paired \textit{t}-test (\texttt{scipy.stats.ttest\_rel}) over the fifteen independent repetitions, after applying the min–max normalization.
\begin{figure}[t] \center
    \includegraphics[width=\textwidth]{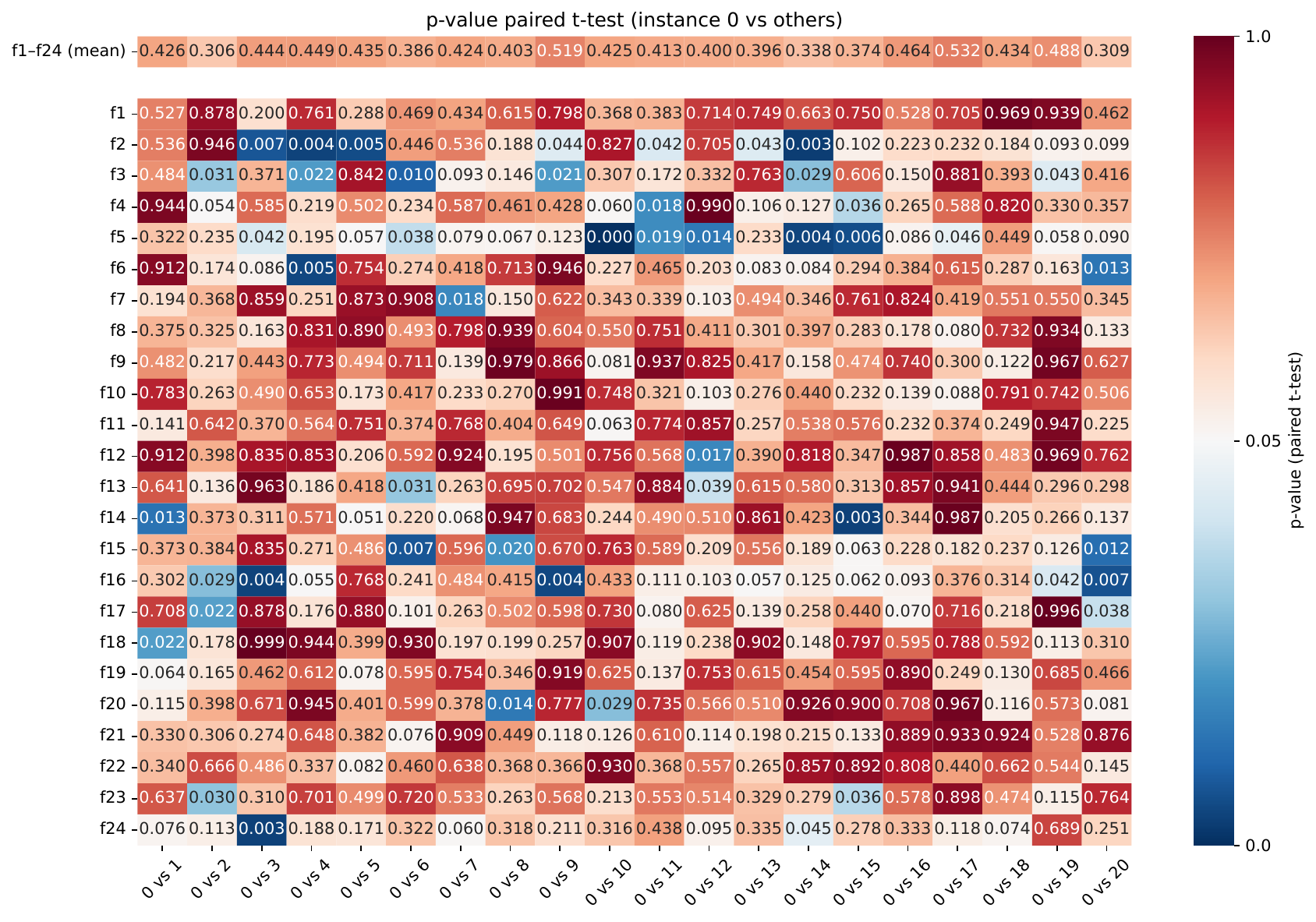}      \caption{Heatmap of p-values from paired \textit{t}-tests comparing the performance of CMA-ES-DS on instance 0 against each of the other 20 instances, for each of the 24 BBOB problems. The comparison is based on the distributions of the average loss of the final batch. The top row shows the average p-value across all functions for each comparison (0 vs 1, ..., 0 vs 20). Below it, each row corresponds to a specific problem (f1–f24). White cells indicate p-values equal to the significance threshold ($\alpha = 0.05$), blue cells denote statistically significant differences (p $<$ 0.05), and red cells indicate no significant difference (p $>$ 0.05). Each p-value was computed using 15 independent repetitions per instance.}
    \label{fig:pvalues}
\end{figure}

To provide an aggregated and intuitive view across all 24 BBOB functions, we visualize the results in a heatmap (Figure~\ref{fig:pvalues}). The top row of the heatmap reports the average p-value for each instance comparison (0 vs 1, 0 vs 2, ..., 0 vs 20), aggregated across all functions. A blank row separates this summary from the detailed view below, where each row corresponds to a specific problem (f1 to f24).

The color coding follows a diverging scheme: p-values equal to the significance threshold ($\alpha = 0.05$) are shown in white, values below the threshold (indicating statistically significant differences) are in blue, and values above the threshold (indicating no significant difference) are shown in red. This visualization helps identify which instance comparisons deviate meaningfully from instance 0 and how often those deviations occur across functions.

As shown in the figure, the top row of average p-values is consistently high across all instance comparisons (all above 0.30), indicating that, on average, instance 0 performs statistically similarly to the other instances. This supports the use of instance 0 as a representative reference throughout the paper. Below the average row, the heatmap reveals a few isolated cases—46 out of 480 total comparisons—where p-values fall below the significance threshold of 0.05. These cases suggest occasional differences in performance for specific problems and instances, but they are sparse and do not indicate a systematic deviation.

\section{CPU Time Analysis Across Hyperparameter Settings}
In this section, we analyze the CPU time required by CMA-ES-DS under different hyperparameter settings. To measure CPU time, we used Python’s \texttt{time.process\_time()} function, which returns the CPU processing time in seconds. This function, available from Python’s \texttt{time} module, measures the actual CPU time consumed by the process, excluding any time during which the process was inactive.
\subsection{Impact of Budget $T$ on CPU Time.}
\begin{figure}[t] \center
    \includegraphics[width=\textwidth]{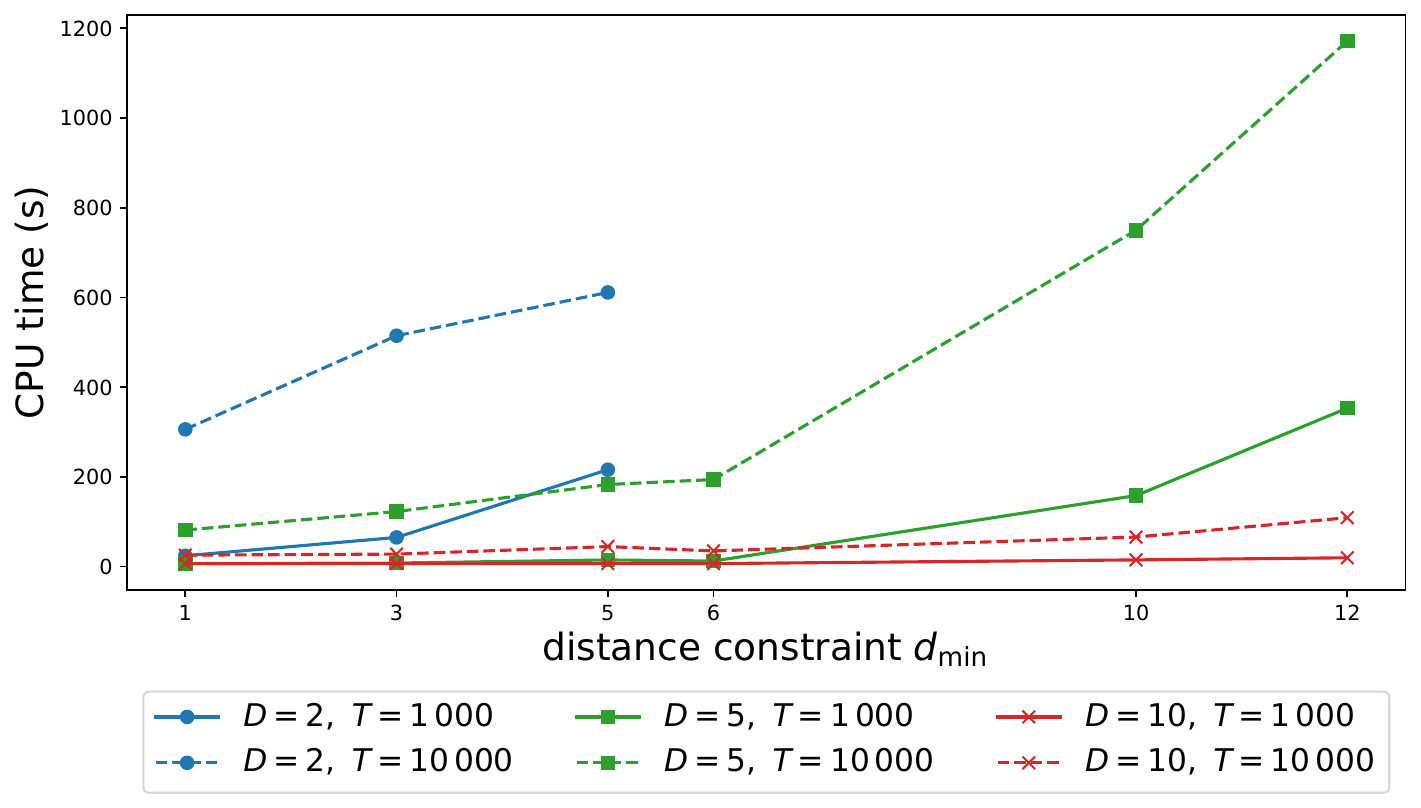}
      \caption{Average CPU time (in seconds) across 24 BBOB functions as a function of the minimum distance constraint. The batch size $k$ is fixed to $k = 5$, while the budget $T$ varies between $T = 1\,000$ and $T = 10\,000$ . Results are averaged over 5 independent runs of CMA-ES-DS. Each line corresponds to a different combination of problem dimension and budget $T$. Lines belonging to the same dimension have the same color and marker.}
    \label{fig:cput}
\end{figure}
Figure~\ref{fig:cput} shows, as expected, that the CPU time increases with the budget $T$. However, the difference in CPU time between the two budget settings becomes smaller as the problem dimension increases. For both budgets and across all dimensions, CPU time grows with the minimum distance constraint $d_{\min}$.
This behavior is due to the increased difficulty of placing $k=5$ valid points in the search space when the dimension $D$ is low or when $d_{\min}$ is large. In such scenarios, our algorithm requires more effort to generate a valid population through cascading checks, where candidate points are iteratively verified for pairwise distance constraints before being evaluated.
For example, when $D=2$, the algorithm struggles significantly at $d_{\min}= 5$, and the CPU time peaks for $T = 10\,000$ in this case. Conversely, for $d_{\min} = 12$, where the smallest analyzed dimension is $D=5$, the CPU time reaches its overall maximum (around 1200 seconds) under budget $T = 10\,000$. On the other hand, for $D=10$, the CPU time remains relatively low: from about 6 seconds (for $T = 1\,000$ and $d_{\min} = 1$) to roughly 120 seconds (for $T = 10\,000$ and $d_{\min} = 12$).
\subsection{Impact of Batch Size $k$ on CPU Time.}
\begin{figure}[t] \center
    \includegraphics[width=\textwidth]{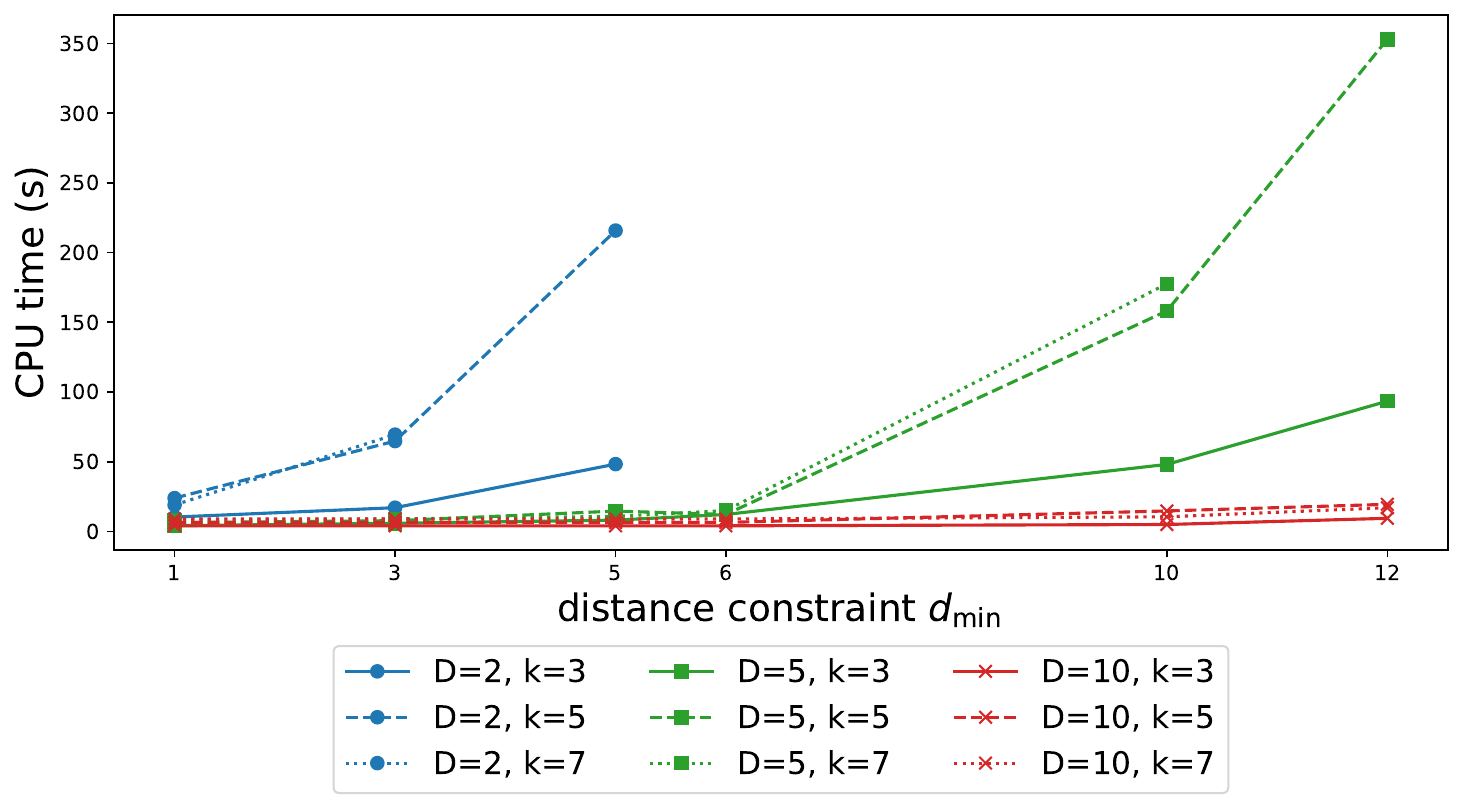}
      \caption{Average CPU time (in seconds) across 24 BBOB functions as a function of the minimum distance constraint.
      The budget  $T$ is fixed to $T = 1\,000$ while the batch size $k$ varies, with $k\in\{3,5,7\}$. Results are averaged over 5 independent runs of CMA-ES-DS. Each line corresponds to a different combination of problem dimension and batch size $k$. Lines belonging to the same dimension have the same color and marker.}
    \label{fig:cpuk}
\end{figure}
Figure~\ref{fig:cpuk} shows that placing $k=7$ valid points in the search space is more challenging than placing $k=3$, which results in generally higher CPU times for larger $k$, especially due to the more intensive cascading checks required to form a valid population. As such, the curve for $k=7$ tends to lie slightly above that for $k=5$, while the difference between $k=3$ and $k=5$ is more pronounced.
Nevertheless, the gap among the curves for different values of $k$ narrows as the problem dimension increases. The most demanding configurations in terms of CPU time are those with $k=7$ and low dimensions. Moreover, in the case of $D = 2$, all runs fail to generate a batch of $k=7$ points at $d_{\min} = 5$, so the curve stops at $d_{\min} = 3$. Similarly, for $D = 5$, the curve stops at $d_{\min} = 10$ due to consistent failures to generate valid batches at $d_{\min} = 12$. In contrast, for $D = 10$, all curves are complete, and the CPU time for $k = 5$ and $k = 7$ becomes nearly identical, indicating that the algorithm handles higher dimensions more efficiently even when larger batch sizes are used since it is generally easier to place a given number of points in a high-dimensional search space than in a low-dimensional one at a pre-defined minimum distance.

\section{Conclusion}
\label{sec:conslusion}

We introduced CMA-ES-DS for the identification of batches of diverse, yet high-quality solutions for single-objective optimization problems with continuous decision variables. Specifically, we addressed the problem of identifying solutions respecting a pairwise distance requirement. 

Our approach demonstrated a strong performance across all evaluated problem settings, with only minor exceptions and particularly strong results for large dimension-to-budget ratios. Key advantages of CMA-ES-DS include: (1) its frequent superiority over other baselines, regardless of the landscape characteristics of the optimized function; (2) its reliability, as it always allows for the extraction of a complete solution batch from its history, unlike the point portfolios generated by other methods; and (3) its robustness, consistently producing point batches of comparable quality across different repetitions on a wide range of considered benchmarks.

In future work, we strive for a deeper understanding of the various trade-offs between leader and alternative solutions. While we evaluated three different subset selection approaches in this study, we have yet to take full advantage of the collected data to investigate the compromises among the different solutions.

We believe our approach to be of high practical importance, and aim to integrate it into state-of-the-art optimization frameworks, to increase its usability for both academic and industrial users.

\begin{acks}
CD acknowledges funding by the European Union (ERC, dynaBBO, 101125586). Views and opinions expressed are however those of the authors only and do not necessarily reflect those of the European Union or the European Research Council. Neither the European Union nor the granting authority can be held responsible for them. 
ER expresses gratitude to Juergen Branke for the initial discussions that inspired some aspects of this project.
We thank the Sorbonne Cluster for Artificial Intelligence (SCAI) for supporting our work through ML's PhD scholarship. 

\end{acks}

\bibliographystyle{ACM-Reference-Format}
\bibliography{main}
\newpage
\onecolumn
\appendix

\section*{Appendix}
\section{Ablation Study on Center Update Strategies and Cascade Dynamics}
\label{appendix:ablation}
In this section, we compare alternative strategies for setting the center of the tabu regions and for handling converged CMA-ES instances in the cascading order. The evaluation was conducted on the most explored setting in our experiments (i.e., dimension  $D = 10$, distance constraint $d_{\min} = 10$, and batch size $k = 5$), as this configuration exhibited the most significant overall performance differences.
We tested three centering strategies:
\begin{itemize}
    \item Using the best individual of the current population (CMA-ES-DS, our final choice in the main paper).
    \item Using the best-so-far solution found by each CMA-ES instance (CMA-ES-DS\_best\_so\_far).
    \item Using the mean of the search distribution (CMA-ES-DS\_mean).
\end{itemize}
We also evaluated an alternative to our current cascading mechanism (CMA-ES-DS\_update\_cascade): when a CMA-ES instance converges, instead of keeping its position in the cascading order unchanged, we tested repositioning it earlier in the cascade based on the temporal order in which instances converge.
\begin{figure}[ht]
    \centering
    \includegraphics[width=\textwidth]{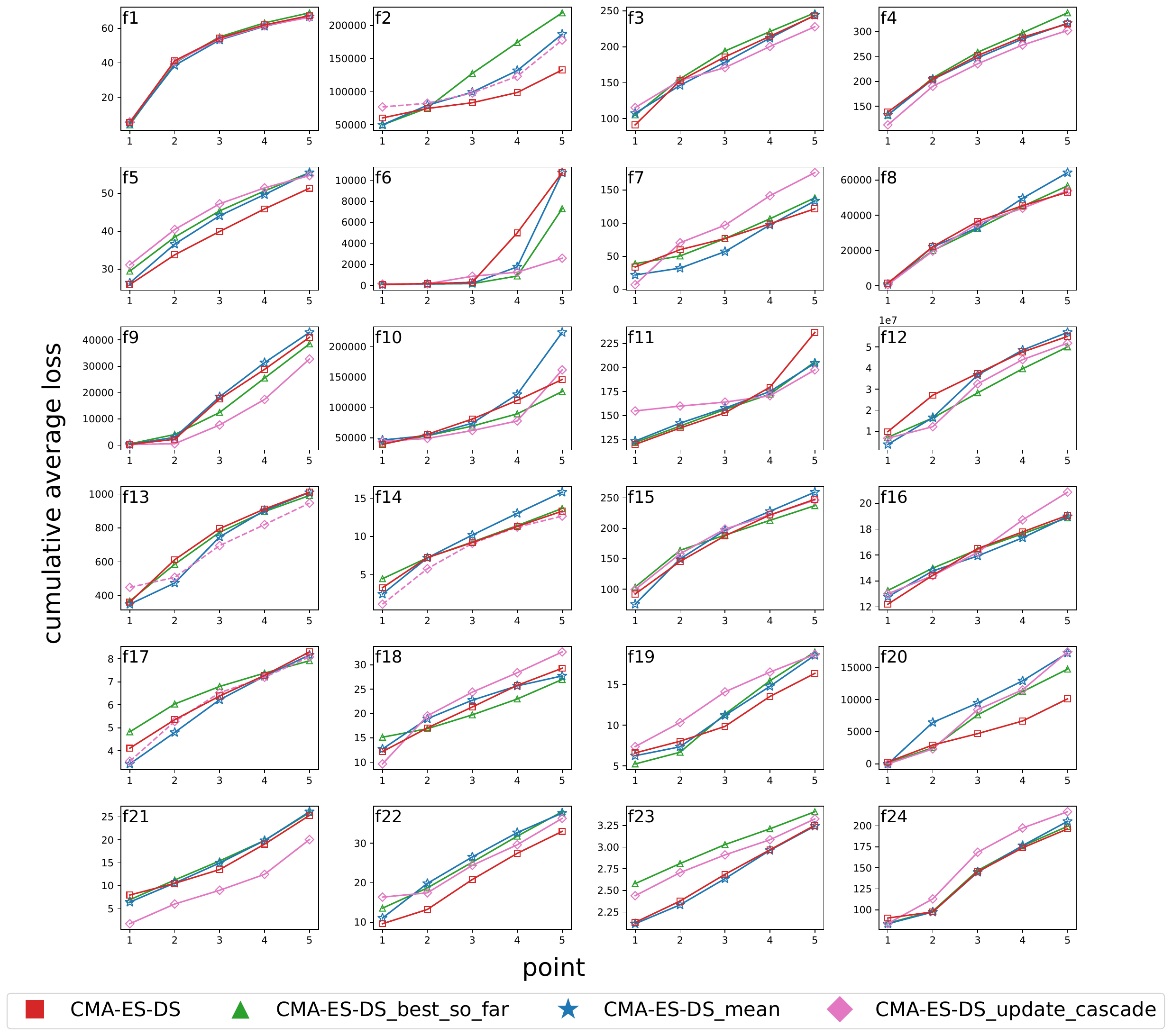}
    \caption{Comparison of the main method for CMA-ES-DS with alternative center update strategies and cascade adjustment.}
    \label{fig:ablation_results}
\end{figure}
Figure~\ref{fig:ablation_results} shows that our current strategy consistently delivers robust and reliable performance across all 24 functions, striking a balance between the first point and the last point found. In contrast, alternatives such as using the mean or best-so-far solution may perform well on specific functions but tend to underperform on others, resulting in less consistent outcomes overall. Similarly, modifying the cascading order upon convergence limits exploration and reduces overall effectiveness and also reliability producing some failures.

\section{Average Loss of the $i$-th Best Point (Non-Cumulative)}
In this section, we provide a complementary analysis to the cumulative average loss shown in the main paper. Specifically, we report the average fitness of the $i$-th best point in the batch, with $i = 1, \dots, 5$. This non-cumulative perspective allows us to assess the quality of each individual point in the batch, without the influence of aggregation effects caused by especially good (leader) or bad points (last alternative point). Moreover, this type of analysis better illustrates how the quality degrades from the best to the worst point in the batch and gives insights into the internal diversity of the selected solutions. It also offers a clearer view of the distribution of fitness values across the batch, and can be more robust to outliers when interpreting performance offering a visualization of the median (the value in point 3).
\begin{figure}[h!] \center
    \includegraphics[width=\textwidth]{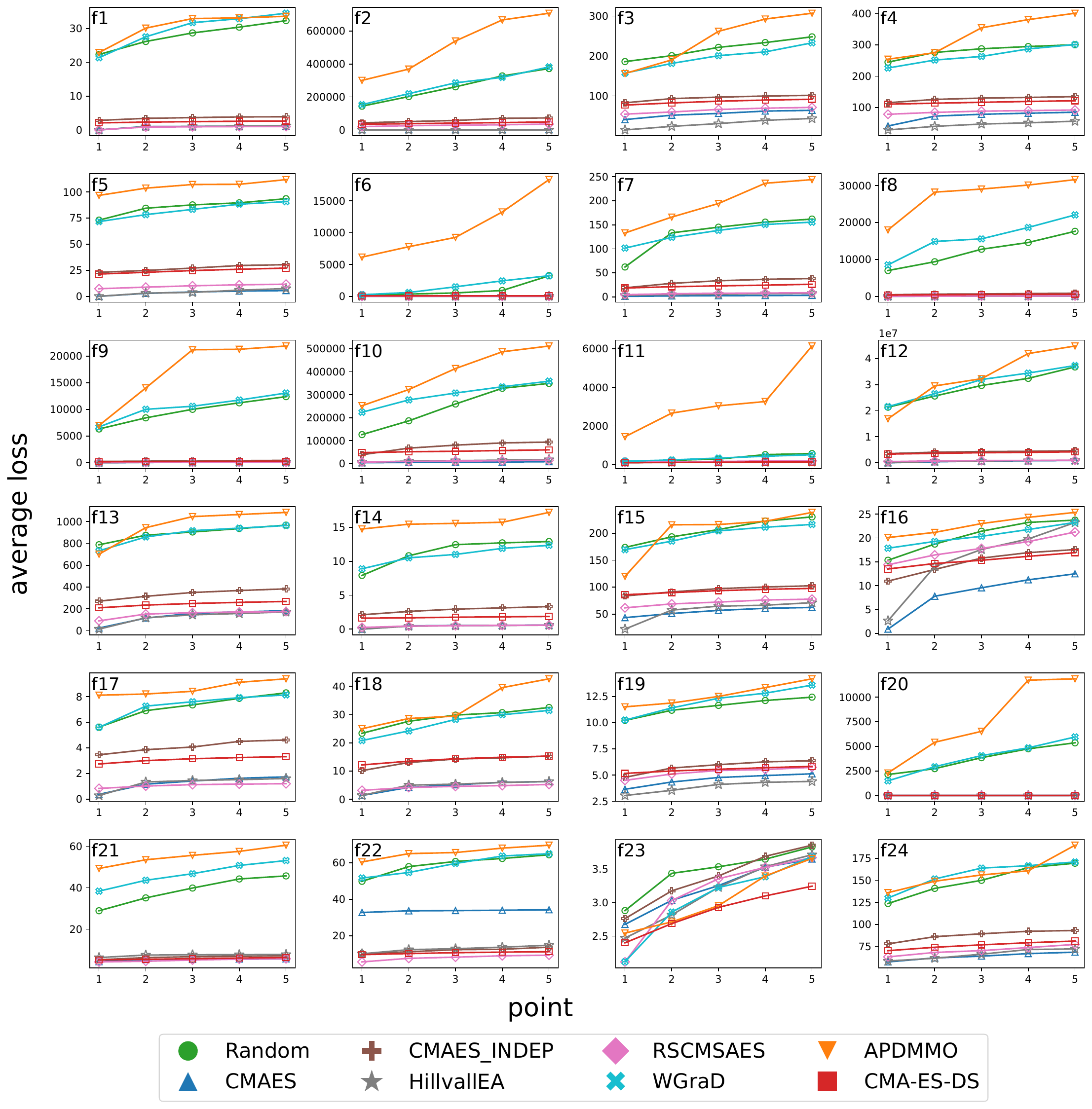}
      \caption{$D = 10$, $T = 1\,000$, $d_{\min} = 1$. Average loss per ranked point in the batch for the 24 BBOB functions.
Each subplot corresponds to one BBOB function. The x-axis indicates the rank of the point in the batch ($x = 1$ for the best, up to $x = 5$), and the y-axis shows the average loss of the $i$-th best point across 5 independent runs. Each curve represents one algorithm.}
    \label{fig:median1}
\end{figure}

\begin{figure}[h!] \center
    \includegraphics[width=\textwidth]{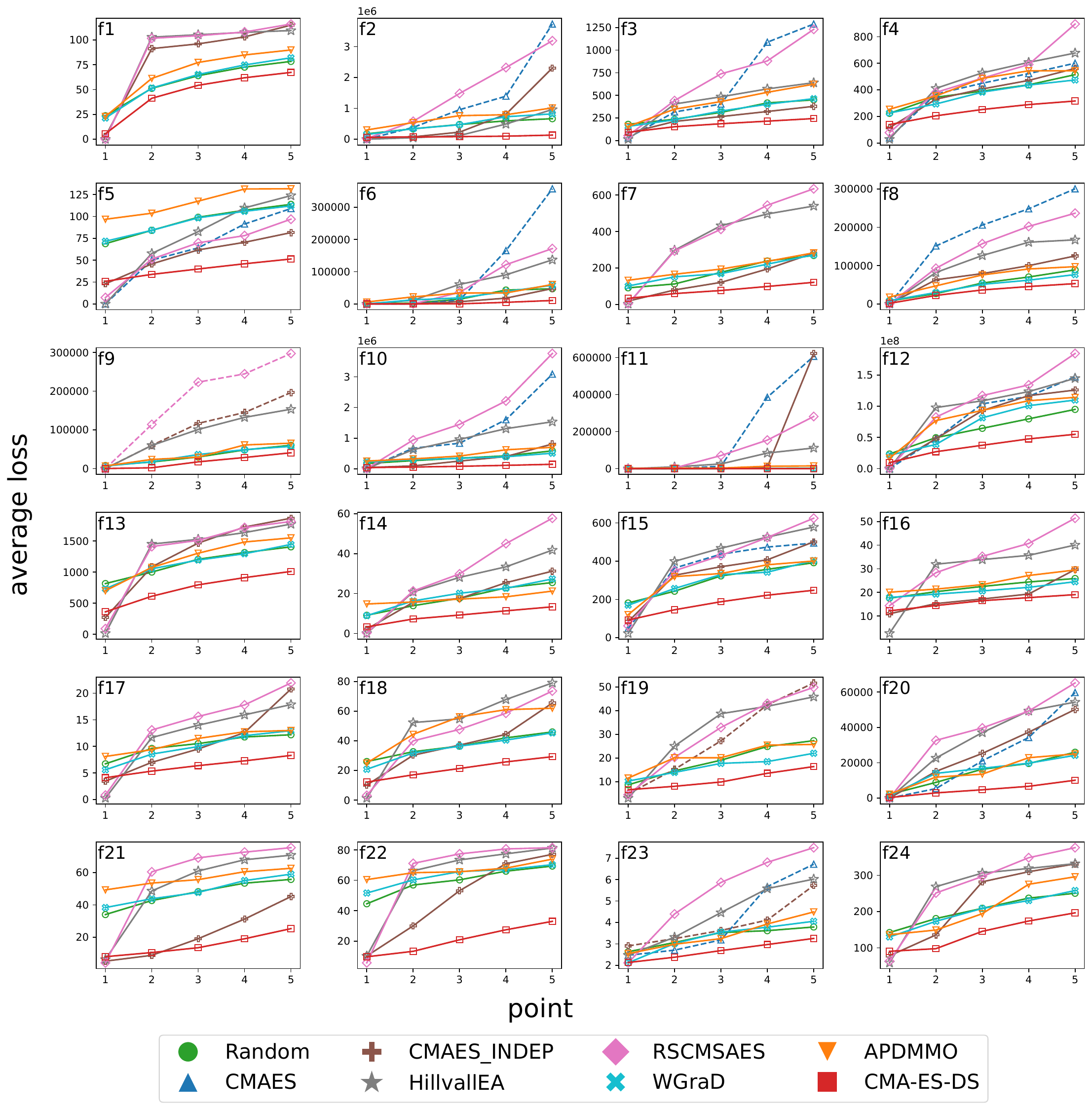}
      \caption{$D = 10$, $T = 1\,000$, $d_{\min} = 10$. Same as Figure ~\ref{fig:median1}, but for $d_{\min} = 10$.}
    \label{fig:median2}
\end{figure}

\begin{figure}[h!] \center
    \includegraphics[width=\textwidth]{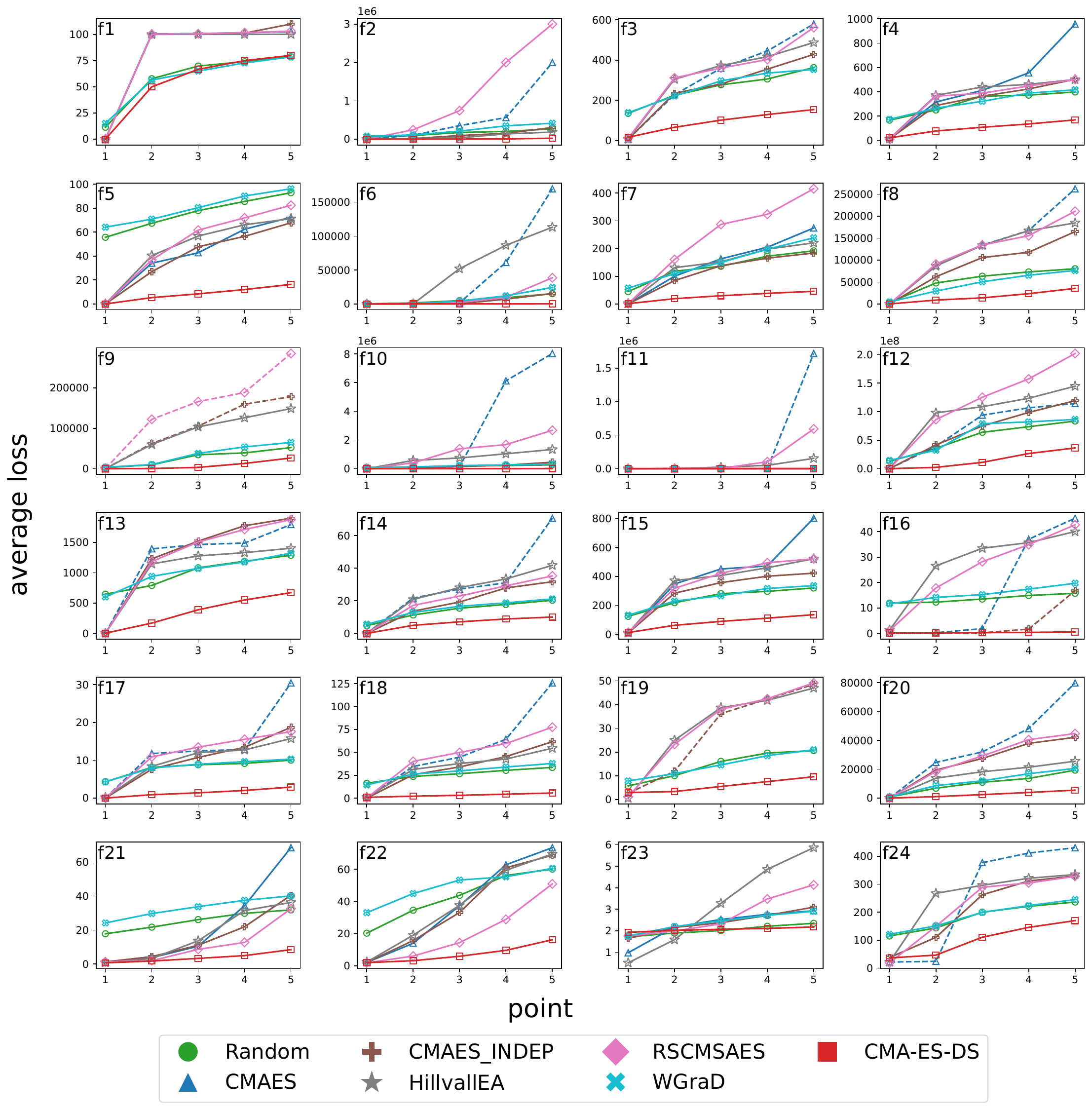}
      \caption{$D = 10$, $T = 10\,000$, $d_{\min} = 10$. Same as Figure ~\ref{fig:median2}, but for initial portfolio size $T = 10\,000$.}
    \label{fig:median3}
\end{figure}
When comparing the non-cumulative plots presented here (Figure~\ref{fig:median1},~\ref{fig:median2}, and~\ref{fig:median3}) with the cumulative average loss plots of Figures 5–7 in the main paper, we observe that CMA-ES-DS appears more competitive in this non-cumulative setting. This difference is particularly evident for smaller budgets (e.g., $T = 1000$), where the first selected point (the batch leader) might be sub-optimal. In the cumulative average plot, a poor leader can negatively impact the cumulative metric, lowering the curve for the entire batch. Another effect, however, can also occur in cumulative plots. This difference is especially noticeable on f1. In the cumulative average plot (Fig, 6-7), CMA-ES-DS starts with a very low loss—indicating an excellent first point—but the curve rises more sharply than those of other algorithms, suggesting lower quality for the remaining points. However, when examining the non-cumulative plot (Fig~\ref{fig:median2}-\ref{fig:median3}), CMA-ES-DS consistently achieves the lowest loss across all ranked batch points. This apparent contradiction is due to the cumulative metric being more sensitive to the deterioration from a strong initial point, whereas the non-cumulative plot reveals that all five batch points are actually of higher quality compared to the competitors. This confirms the value of non-cumulative plots for a clearer assessment of batch composition quality.
\section{Additional Results for Various Settings}
\begin{figure*}[h!] \center
    \includegraphics[width=\textwidth]{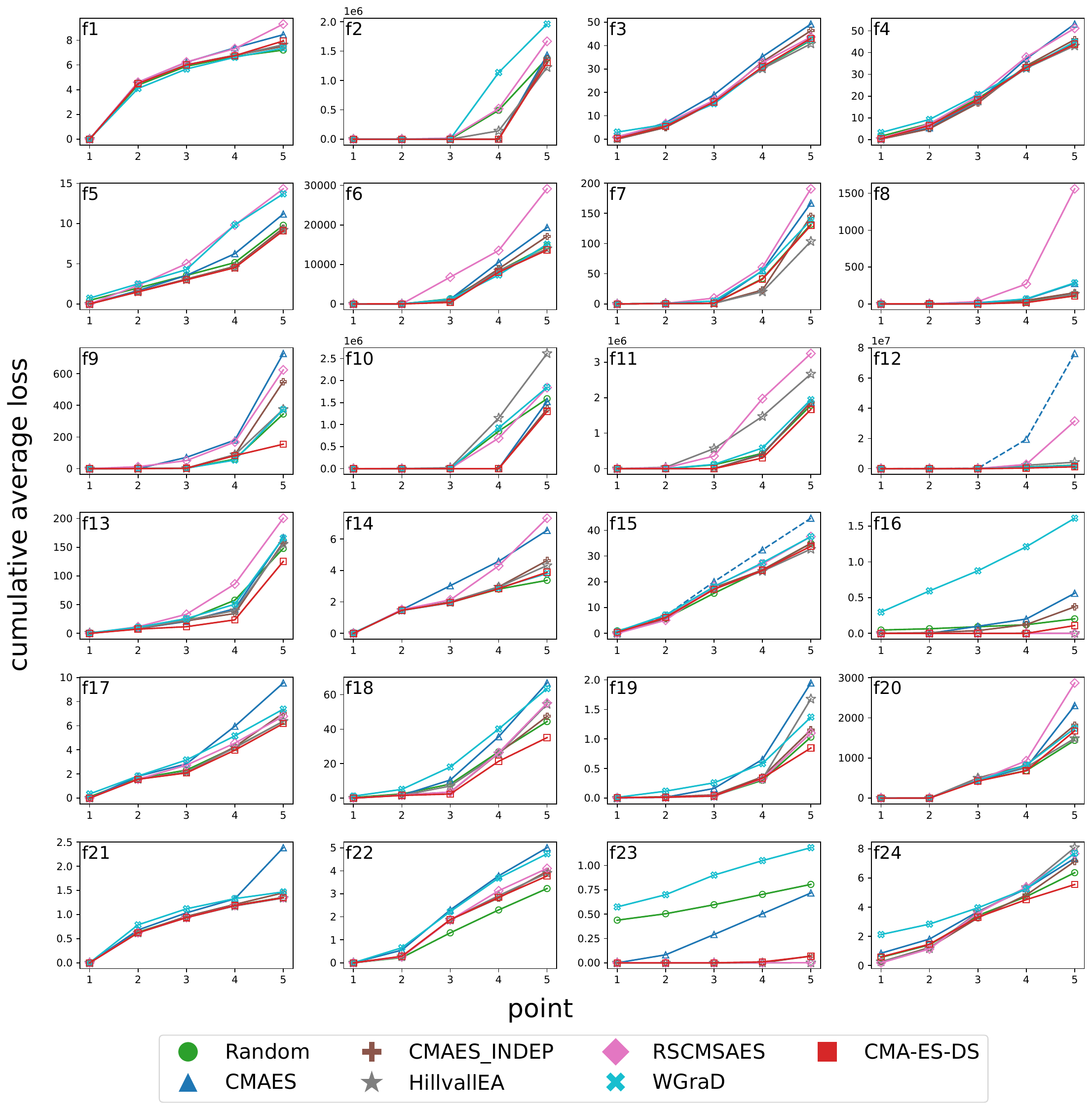}
      \caption{$D = 2$, $T = 10\,000$, $d_{\min} = 3$. Cumulative average loss across the $k = 5$ points in the batch for the 24 BBOB functions. Each subplot corresponds to a specific BBOB function, and the x-axis represents the batch points ($x=1$ to $x=5$). The y-axis shows the cumulative average loss, where each curve represents the mean performance of an algorithm over 5 independent runs.}
    \label{fig:D210Kd3}
\end{figure*}
\clearpage
\begin{figure*}[h!] \center
    \includegraphics[width=\textwidth]{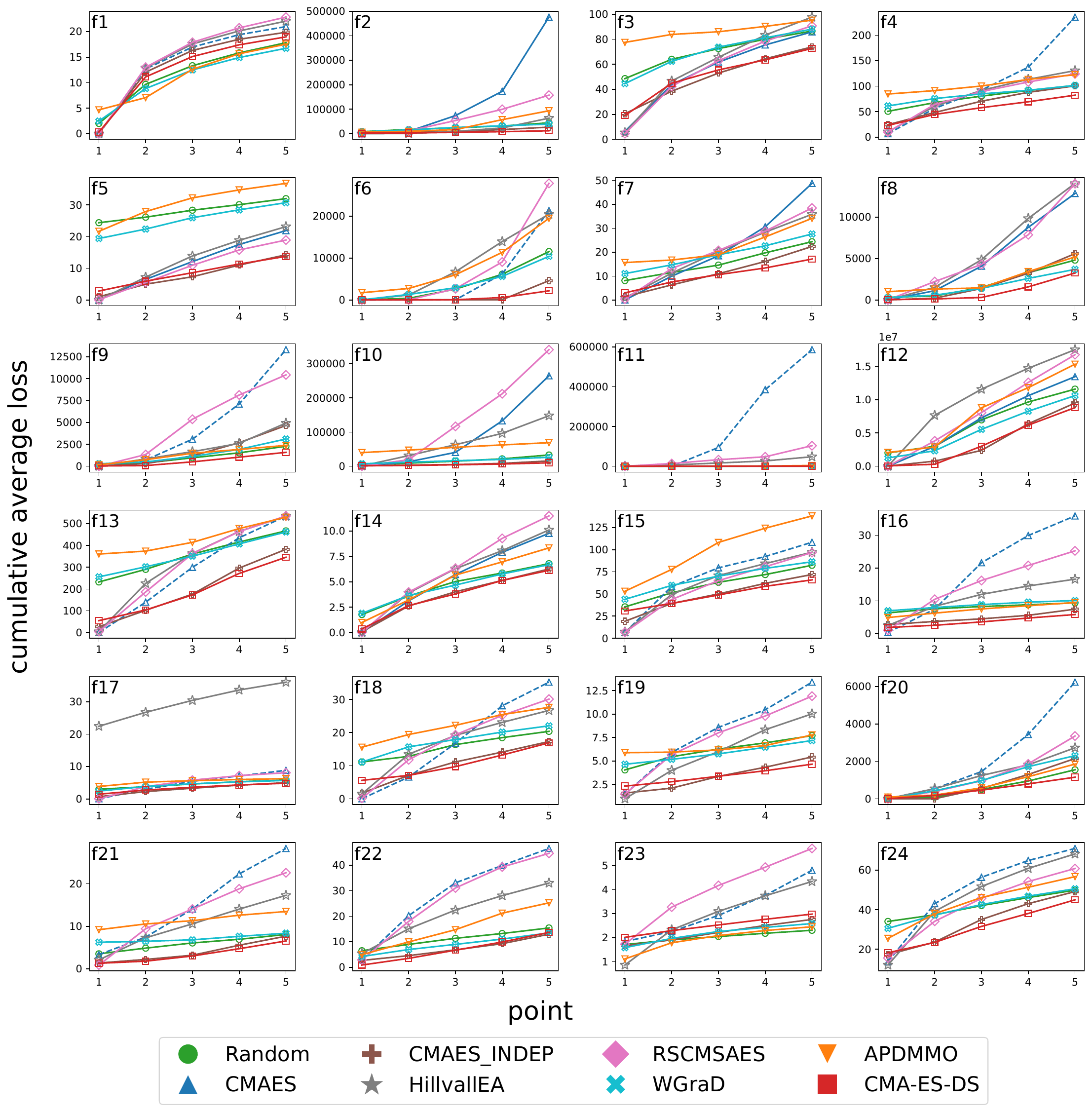}
      \caption{$D = 5$, $T = 1\,000$, $d_{\min} = 5$. Cumulative average loss across the $k = 5$ points in the batch for the 24 BBOB functions. Each subplot corresponds to a specific BBOB function, and the x-axis represents the batch points ($x=1$ to $x=5$). The y-axis shows the cumulative average loss, where each curve represents the mean performance of an algorithm over 5 independent runs.}
    \label{fig:D51Kd5}
\end{figure*}

\begin{figure*}[htbp]
    \centering
    \begin{subfigure}[b]{0.45\textwidth}
        \centering
        \includegraphics[width=\textwidth]{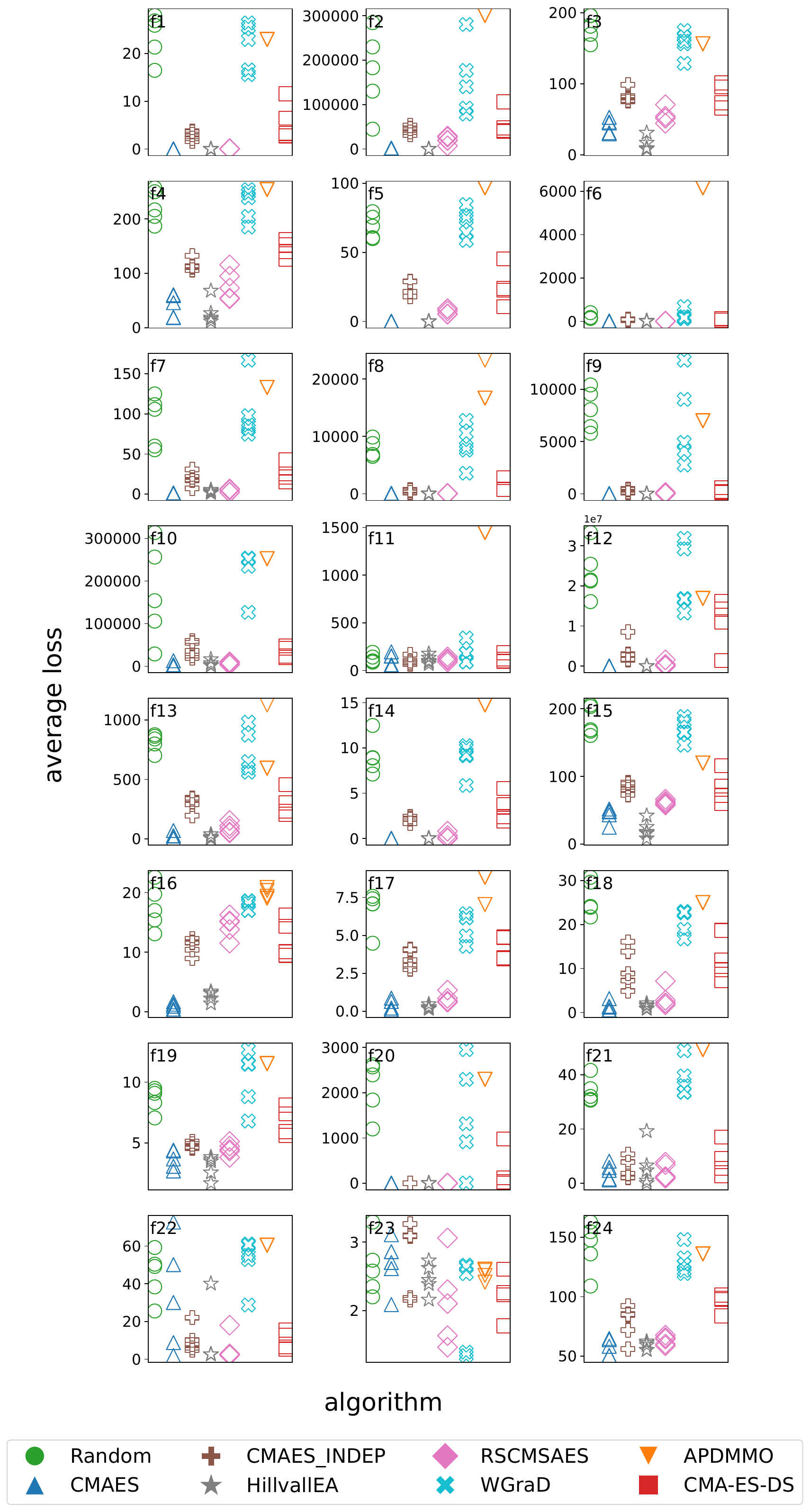}
        \caption{Scatter plot of the leader's loss values across 5 independent runs for each algorithm. Each figure represents the leader's loss in a single run.}
        \label{fig:img1}
    \end{subfigure}
    \hspace{0.05\textwidth}
    \begin{subfigure}[b]{0.45\textwidth}
        \centering
        \includegraphics[width=\textwidth]{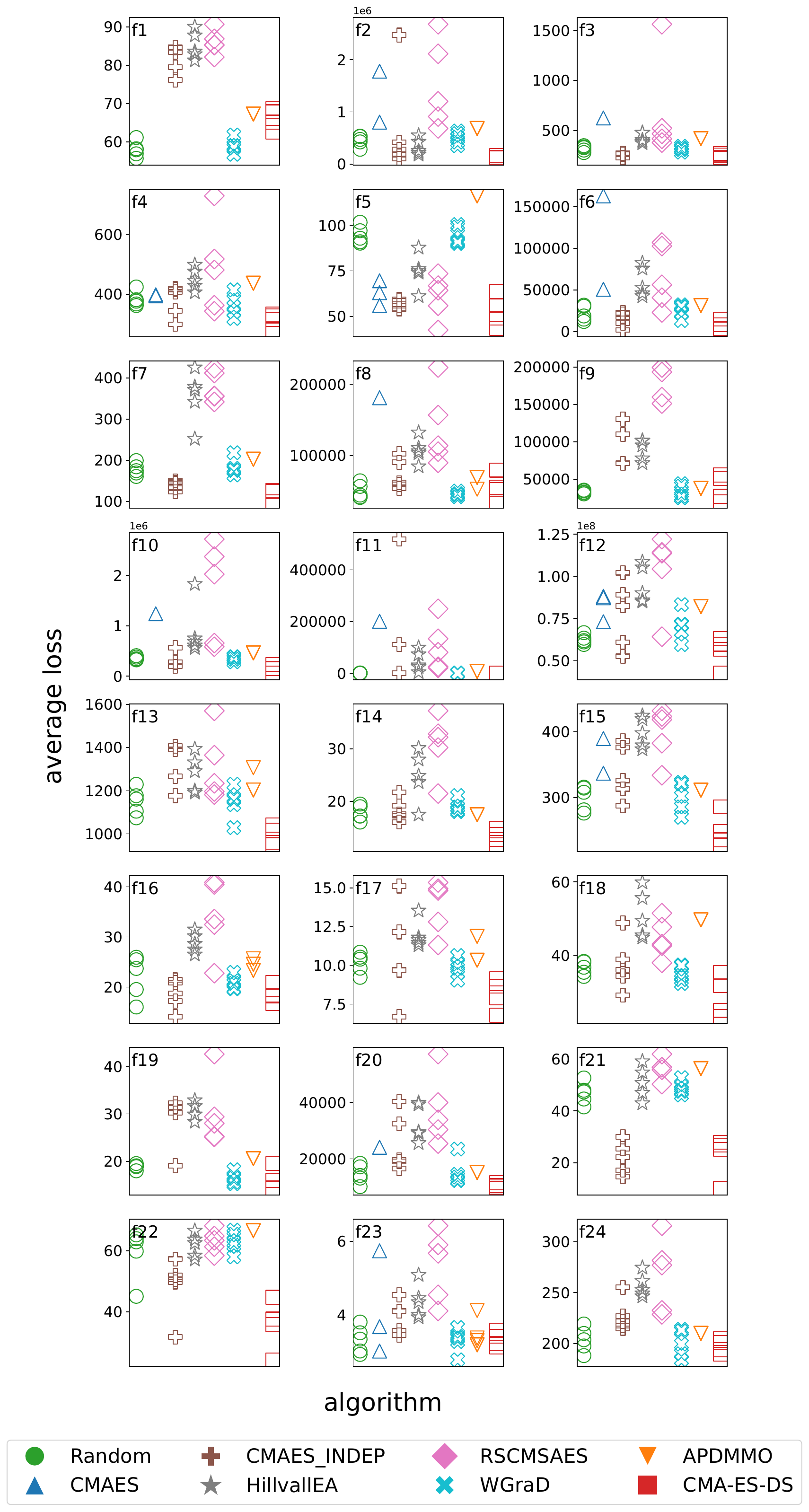}
        \caption{Scatter plot of the average loss over the batch of $k = 5$ points for each algorithm. Each figure represents the average batch loss from a single run.}
        \label{fig:img2}
    \end{subfigure}
    \caption{$D = 10$, $T = 1\,000$, $d_{\min} = 10$.}
    \label{fig:variance}
\end{figure*}

\end{document}